\documentclass[journal]{IEEEtran}

\usepackage{xcolor,soul,framed} 

\colorlet{shadecolor}{yellow}
\usepackage[pdftex]{graphicx}
\graphicspath{{../pdf/}{../jpeg/}}
\DeclareGraphicsExtensions{.pdf,.jpeg,.png}

\usepackage[cmex10]{amsmath}
\usepackage{array}
\usepackage{mdwmath}
\usepackage{mdwtab}
\usepackage{eqparbox}
\usepackage{url}

\usepackage{cite}
\usepackage{amsmath,amssymb,amsfonts}
\usepackage{algorithm}          
\usepackage{algorithmicx}
\usepackage{algpseudocode}

\usepackage{multirow}
\usepackage{multicol}

\makeatletter
\algnewcommand{\LeftComment}[1]{\Statex /* \textit{#1} */}

\usepackage{forloop}
\newcounter{ct}
\newcommand{\indentOnce}[1]{\forloop{ct}{0}{\value{ct} < #1}{\hspace{\algorithmicindent}}}
\newcommand{\indentComment}[1]{\Statex\indentOnce{#1}}

\newlength{\trianglerightwidth}
\algnewcommand{\LineCommentCont}[1]{\Statex \hskip\ALG@thistlm%
  \parbox[t]{\dimexpr\linewidth-\ALG@thistlm}
{\leftskip=\algorithmicindent
  \hangindent=\algorithmicindent 
  \hangafter=1%
  \strut\makebox[\algorithmicindent][c]/*\textit{#1}*/\strut}
  } 
\makeatother

\usepackage{caption}

\usepackage{multirow}
\usepackage{multicol}

\usepackage{hyperref}
\hypersetup{
    colorlinks=true,
    citecolor=black,
    linkcolor=black,
    filecolor=black,      
    urlcolor=black,
}

\begin{document}
\bstctlcite{IEEEexample:BSTcontrol}
    \title{Online Multi-Object Tracking Framework with the GMPHD Filter and Occlusion Group Management}
  \author{Young-min Song, Kwangjin Yoon, Young-Chul Yoon, Kin-Choong Yow, and Moongu Jeon*

  \thanks{Y. Song, K. Yoon, and M. Jeon are with the School of Electrical Engineering and Computer Science, Gwangju Institute of Science and Technology, Gwangju 61005, South Korea (e-mail: sym@gist.ac.kr; yoon28@gist.ac.kr; mgjeon@gist.ac.kr).}
  \thanks{Y.-C. Yoon is with Convergence Biz. Development Office, LG Electronics, Seoul 07796, South Korea (email : youngchul.yoon@lge.com).}%
  \thanks{K.-C. Yow is with the Faculty of Engineering and Applied Science, University of Regina, Regina S4S 0A2, Canada (e-mail: Kin-Choong.Yow@uregina.ca).}
}


\maketitle

\begin{abstract}
In this paper, we propose an efficient online multi-object tracking framework based on the Gaussian mixture probability hypothesis density (GMPHD) filter and occlusion group management scheme where the GMPHD filter utilizes hierarchical data association to reduce the false negatives caused by miss detection. The hierarchical data association consists of two steps: detection-to-track and track-to-track associations, which can recover the lost tracks and their switched IDs. 
In addition, the proposed framework is equipped with an object grouping management scheme which handles occlusion problems with two main parts. The first part is ``track merging" which can merge the false positive tracks caused by false positive detections from occlusions, where the false positive tracks are usually occluded with a measure. The measure is the occlusion ratio between visual objects, sum-of-intersection-over-area (SIOA) we defined instead of intersection-over-union (IOU) metric. 
The second part is ``occlusion group energy minimization (OGEM)" which prevents the occluded true positive tracks from false ``track merging". We define each group of the occluded objects as an energy function and find an optimal hypothesis which makes the energy minimal.
We evaluate the proposed tracker in benchmark datasets such as MOT15 and MOT17 which are built for multi-person tracking. An ablation study in training dataset shows that not only ``track merging" and ``OGEM" complement each other but also the proposed tracking method has more robust performance and less sensitive to parameters than baseline methods. Also, SIOA works better than IOU for various sizes of false positives. Experimental results show that the proposed tracker efficiently handles occlusion situations and achieves competitive performance compared to the state-of-the-art methods. Especially, our method shows the best multi-object tracking accuracy among the online and real-time executable methods. 
\end{abstract}


%
\IEEEpeerreviewmaketitle


\section{Introduction}

\IEEEPARstart{M}{ulti-object} Tracking (MOT) has become one of key techniques for intelligent video surveillance~\cite{MOT15,MOT16} and autonomous vehicle systems~\cite{kitti} in the last decade. 

In view of the processing pipelines, many state-of-the-art MOT methods~\cite{prev1,prev2,sort,bae2, ham, fu1, mdp, tbss, scea, eamtt, rmot, gscr, bae1, cnntcm, siameseCNN, nomt, mhtdam, elp, jpdam, moticon, segtrack, cem, smot, dpnms, motdt, mtdf, dman, amadm, gmphdn1tr, gmphdkcf, gmphd2012, ehaf, fwt, jcc, tlmht, edmt, iou, mhtlstm} have exploited the tracking-by-detection paradigm. This phenomenon has been standing out while deep neural networks based detectors such as FRCNN~\cite{frcnn}, SDP~\cite{sdp}, and EB\cite{eb} have shown breakthrough in object classification and detection.

Besides, MOT algorithms are categorized into two approaches: offline and online processes. The most different point between two approaches is that whereas the offline process can see the whole time sequences at once, the online process can see only the frames from initial time $1$ to current processing time $k$. In other words, from the system user's perspective, whereas the offline method is suitable for post-processing, the online process for real-time application.

Thus, many offline methods~\cite{jcc,mhtdam,jpdam,nomt,dpnms,cem} take advantage of the global optimization models. 
~\cite{dpnms,nomt,jcc} exploit graphical models to solve MOT task.
Pirsiavash~\textit{et al.}\cite{dpnms} designed a min-cost flow network where the nodes and the directed edges indicating observations and tracklets' hypotheses, respectively form a directed acyclic graph (DAG). The DAG's shortest (min-cost) path can be found with Dijkstra's algorithm.
Choi~\textit{et al.}\cite{nomt} divided the tracking problem into subgraphs and solved each subgraph as conditional random field inference in parallel. 
Keuper~\textit{et al.}\cite{jcc} applied vision-based perspective to the proposed graph optimization model. Feature points' trajectories and bounding boxes build low-level and high-level graph models, respectively, and then, they find the optimal association results between the two levels graph models.
Rezatofighi~\textit{et al.}\cite{jpdam} and Kim~\textit{et al.}\cite{mhtdam} considered all possible hypotheses for data association. Because it involves the exponentially increasing complexity with a tree structure, \cite{jpdam} assumed $m$-best solutions and \cite{mhtdam} pruned out invalid hypotheses using their own rule. 
Besides, Milan~\textit{et al.}\cite{cem} proposed a sophisticated energy minimization technique considering detection, appearance, dynamic model, mutual exclusion, and target persistence for MOT task in video.
Those offline methods have strength to generate the accurate and refined tracking results but is not suitable for practical real-time application.

On the other hand, since the online approach cannot apply the global optimization models, intensive motion analysis and appearance feature learning have been popularly utilized with a hierarchical data association framework and the online Bayesian model~\cite{rmot,scea,bae1,bae2,ham,motdt,gmphdkcf,fu1}.
Yoon~\textit{et al.}\cite{rmot} proposed a relative motion analysis between all objects in a frame, and then improved the work~\cite{rmot} by adding the cost optimization function using context constraints in~\cite{scea}.
Bae~\textit{et al.}\cite{bae1} exploited the incremental linear discriminant analysis (LDA) for appearance learning and presented a tracklet confidence based data association framework. Also, in~\cite{bae2}, they improved their previous work~\cite{bae1} by using the deep neural network (DNN) based appearance learning instead of the incremental LDA. As we addressed in the previous paragraph, DNN has given breakthrough in appearance learning i.e., object classification and detection. 
So, some online MOT algorithms have focused on how to adopt deep appearance learning into their tracking frameworks. 
Yoon~\textit{et al.}\cite{ham} exploited the siamese convolutional neural networks (CNN)~\cite{siamese} to train appearance model. They train the deep appearance networks selectively where only the detection responses matched with high confidence between the historical object queues in the recent few frames. Then, they combine the trained networks to a simple Bayesian tracking model with the Kalman filter. Chen~\textit{et al.}\cite{motdt} employed a re-identification (Re-ID) model~\cite{reid} to their tracking framework. They measure the similarity between detection and track by calculating the distance between Re-ID feature vectors of them. Then, they associate the pairs of detections and tracks which make the sum of the distances minimal.
Both approaches~\cite{ham,motdt} proposed online Bayesian tracking models with conventional DNN models to measure the similarity between the visual objects.
Those online MOT methods have proposed successful solutions with excellent tracking accuracy but their intensive analysis and learning processes take heavy computing resource and time. Also, even if they just employ conventional DNN models through state-of-the-art GPU processing technique, the requirement for a lot of computing resource is inevitable and it makes the trackers difficult to achieve real-time speed.

Recently, the closed-form implementations~\cite{smcphd,gmphd} of the probability hypothesis density (PHD) filtering have been employed as an emerging theory for many online MOT methods~\cite{gmphdn1tr,gmphdkcf,gmphd2012,fu1,eamtt,mtdf,prev1,prev2}. That is because
Vo~\textit{et al.}\cite{smcphd,gmphd} provided not only theoretically optimal approach to the online multi-target Bayes filtering but also approximate the original PHD recursions involving multiple integrals, which alleviate the computational intractability. Moreover, the PHD filter was originally designed for multi-target tracking in radar/sonar systems which receive uncountable false positive observations, i.e., clutters. So it is robust to deal with false positives errors but weak to handle false negatives.
V. Eiselein~\textit{et al.}\cite{gmphd2012} combined the feature-based label tree to the Gaussian mixture PHD (GMPHD) filter, which use visual features to help the GMPHD filter work sensibly in video data system. Song~\text{et al.}\cite{prev1} extended the GMPHD filter based tracking with the two-stage hierarchical data association strategy and use simple motion estimation and appearance matching to recover lost tracks.
T. Kutschbach~\textit{et al.}\cite{gmphdkcf} joined the GMPHD filter with the kernelized correlation filters (KCF)~\cite{kcf} for online appearance update to overcome occlusion.
Z. Fu~\textit{et al.}\cite{fu1} adopted an adaptive gating technique and an online group-structured dictionary (appearance) learning strategy into the GMPHD filter. They make the GMPHD filter be sophisticated and fit to video based MOT.
Besides, various tracking methods~\cite{gmphdn1tr,eamtt,mtdf} utilizing the PHD filters have been proposed.

These latest MOT research trends motivate our work in terms of the three main contributions.
Also, it reminds us of the requirements for the practical MOT applications.
Thus, in this paper, we propose an online multi-object tracking framework to resolve the practical tracking problems which are based on occlusion and the characteristics of video data system.
First, we exploit the GMPHD filter for online MOT. To efficiently change the GMPHD filter's original domain, we define the tracking problems by miss detections in video data system. To deal with track loss by miss detection, we design a GMPHD filtering theory based hierarchical data association (HDA) strategy.
Second, we assume that most of tracking problems are caused by occlusion in video data system. The occlusion between false positive tracks can cause ID-switch and the false positives, and the real occlusion between objects can make fragmented and miss tracks by miss detections. To handle these tracking problems, we propose a novel occlusion handling technique combined with HDA which is based on GMPHD filter tracking framework.
Third, we consider that the proposed tracking framework should be implemented to run with real-time speed.
That is because visual surveillance systems with higher intelligence require more immediate responses to the users with real-time speed. Also, immediate responses can help the systems' user and the machines to react abnormal situation rapidly.
Finally, we evaluate the proposed method on the popular benchmark dataset. Our method shows the competitive performance against state-of-the-art methods in terms of ``tracking accuracy versus speed".

Our main contributions are described as follows:

1) To apply the GMPHD filter into video data system, we extended the conventional GMPHD filter based tracking process with a hierarchical data association (HDA) strategy. Also, we revised the equations of the GMPHD filter as a new cost function for HDA. HDA consists of detection-to-track association (D2TA) and track-to-track association (T2TA). Each cost matrix of each association stage is solved by the Hungarian method with the linear complexity $O(n^3)$ (assignment problem). These D2TA and T2TA recovers lost tracks, while preserving real-time speed.

2) To handle occlusion in video-based tracking system, we devised ``tracking merging" and ``occlusion group energy minimization (OGEM)" which complement each other. ``Tracking merging" relieves false positive tracks and ``OGEM" recovers false ``track merging" by using the occluded objects' group energy minimization. ``Tracking merging" runs in tracking-level so is different to detection-level merging such as non-maximum-suppression. To measure overlapping ratio between occluded objects, we devise a new metric named as sum-of-intersection-over-area (SIOA). We use the SIOA metric instead of intersection-over-union (IOU) which is an extensively used metric.
For ``OGEM", we devise a new energy function to find the optimal state having the minimum energy in a group of occluded objects. ``Tracking merging" and ``OGEM" follow D2TA and T2TA, respectively. We name both techniques as occlusion group management (OGM).


3) Consequently, we propose an online multi-object tracking framework with the GMPHD filter and occlusion group management (GMPHD-OGM). In view of optimization techniques, the first and second contribution locally optimize tracking process which are the minimization of the association cost matrix and the occlusion group energy. 
We evaluate the proposed tracking framework on MOT15~\cite{MOT15} and MOT17~\cite{MOT16} benchmarks. The ablation study on training set shows that our method is more robust than the given baselines. The qualitative and quantitative evaluation results shows that GMPHD-OGM efficiently handle the defined tracking problems by occlusion. Moreover, the proposed method achieves competitive tracking performance against state-of-the-art online MOT algorithms in terms of CLEAR-MOT metrics~\cite{clearmot}. 


The related works are described in Section~\ref{sec:relatedwork}.
In Section~\ref{Framework} and~\ref{Contribution}, we introduce the GMPHD filter based tracking framework with HDA and OGM in detail, respectively.
In Section~\ref{Experiments}, our method is evaluated compared to baseline methods and state-of-the-art methods on the popular benchmarks MOT15~\cite{MOT15} and MOT17~\cite{MOT16}. We conclude this paper with future work in Section~\ref{Conclusion}.
Some preliminary results of this work was presented in Song~\textit{et al.}\cite{prev1,prev2}. 

\begin{figure*}
\centering
\includegraphics[width=18cm]{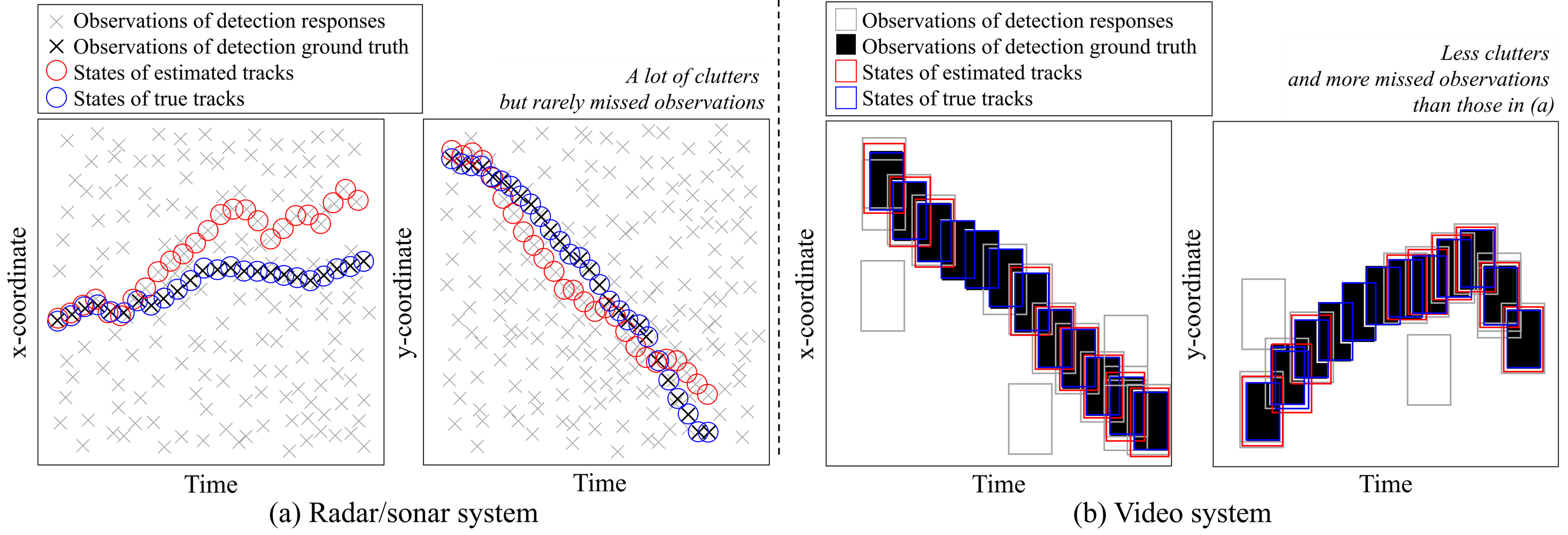}
\caption{Comparison between (a) radar/sonar and (b) video system in terms of input and output, i.e., observations (detection) and states (tracking). 
The radar/sonar sensors receive a lot of clutters (false positive error) but rarely miss objects (false negative error), whereas the detector in video data tends to receive a few clutters around the objects and misses more objects than the radar/sonar senors do.}
\label{fig:domains}
\end{figure*}

\section{Related Works}
\label{sec:relatedwork}

Our proposed tracking framework is influenced from the PHD filter based online multi-object tracking, and grouping approach (topology and relative motion analysis).

The PHD filter\cite{mahler,gmphd,smcphd} was originally designed to deal with radar/sonar data based multi-object tracking (MOT) systems. Mahler~\textit{et al.}\cite{mahler} proposed a recursive Bayes filter equations for the PHD filter which optimizes MOT process in radar/sonar systems with the random-finite set (RFS) of state and observations. Following this PHD filtering theory, Vo~\textit{et al.}\cite{gmphd} implemented governing equations by using the Gaussian mixture model as closed-form recursions, named as the Gaussian mixture probability hypothesis density (GMPHD) filter.
In the original domains the tracking algorithm should estimate true tracks (states) from a lot of observations as shown in Figure~\ref{fig:domains}-(a). Whereas the radar/sonar sensors receive massive false positive but rarely missed observations, visual object detectors generate much less false positive and more missed observations than the radar/sonar sensors does as shown in Figure~\ref{fig:domains}-(b). 
Thus, the GMPHD filter is efficient dealing with the false positive observations, but needs to be extended and improved by additional techniques for MOT in video data system.

As demand increases on online and real-time tracker in video-based tracking system, the PHD filter have been an emerging tracking model, recently.
Song~\text{et al.}\cite{prev1} extended the GMPHD filter based tracking with the two-stage hierarchical data association strategy to recover fragmented and lost tracks. They defined the affinity in the track-to-track association step by using tracks' linear motion and color histogram appearance. This approach is an intuitive implementation of the GMPHD filter to handle tracking problems, but cannot correct the false associations already made in the detection-to-track association.
T. Kutschbach~\textit{et al.}\cite{gmphdkcf} added the kernelized correlation filters (KCF)~\cite{kcf} for online appearance update to overcome occlusion with the naive GMPHD filtering process. They showed a robust online appearance learning to re-find the IDs of the lost tracks. However, updating appearance information of all objects at every frame requires heavy computing resources.
R. Sanchez-Matilla~\textit{et al.}\cite{eamtt} proposed a detection confidence based MOT model with the PHD filter. Strong (high confidence) detections initiate and propagate tracks but weak (low confidence) detections only propagate existing tracks. This strategy works well when the detection results are reliable. However, the tracking performance is dependent on the detection performance, and especially weak to long-term missed detections.
Z. Fu~\textit{et al.}\cite{fu1} adopted an adaptive gating technique and an online group-structured dictionary (appearance) learning strategy into the GMPHD filter. They made the GMPHD filter have a sophisticated tracking process and fit to video based MOT.

Grouping approach e.g., relative motion and topological model, already have been exploited in \cite{rmot,scea}. The key difference between their methods and ours is that \cite{rmot,scea} consider the relations between all objects in a scene but we only consider topological information in the group of occluded objects.
Grouping only the occluded objects exclude trivial solutions (associations) which focuses on solving sub-problems and reduces computing time.

\begin{figure*}
\centering
\includegraphics[width=18cm]{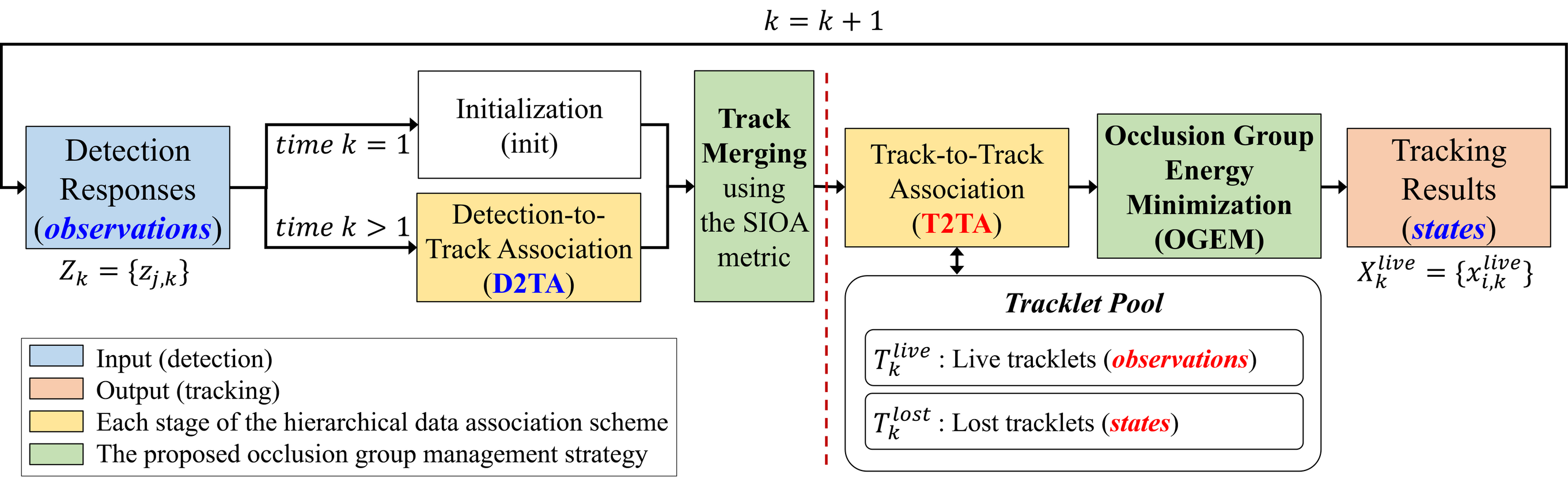}
\caption{Flow chart of the proposed online multi-object tracking framework. The red dotted line divides the proposed hierarchical data association into two stages. Each stage and its states and observations are marked as \textbf{\textcolor{blue}{blue}} and \textbf{\textcolor{red}{red}}, i.e., D2TA and T2TA, respectively. The key components of this chart, such as init, D2TA, Merge, T2TA, OGEM, live and lost tracklets are used in Figure~\ref{fig:state_machine} and Figure~\ref{fig:qual_results}, also.}
\label{fig:framework}
\end{figure*}

\section{Proposed Online Multi-Object Tracking Framework}
\label{Framework}

In this section, we briefly introduce the general tracking process of the Gaussian mixture probability hypothesis density (GMPHD) filter in Subsection \ref{subsec:gmphd}. In \ref{hda}, we address how to extend the GMPHD filter with the hierarchical data association strategy in video-based online MOT systems. 
\subsection{The GMPHD Filter}
\label{subsec:gmphd}
The Gaussian mixture model (GMM) of the GMPHD filter includes means, covariances, and weights which are propagated at every time stamp as follows; \textit{Initialization}, \textit{Prediction}, \textit{Update}, and \textit{Pruning} steps.
We employ this basic process of the GMPHD filter but revise fit to the video-based MOT system.

\begin{align}
&X_k = \{x^1_k,...,x^{I_{k}}_k\}, \label{eq1}\\
&Z_k = \{z^1_k,...,z^{J_{k}}_k\}, \label{eq2}
\end{align}
where $X_k$ and $I_{k}$ denote a set of objects' states and the number of them at time $k$, respectively. A state vector $x_k$ is composed of $(c_x,c_y,v_x,v_y)$, where $c_x$, $c_y$, $v_x$, and $v_y$ indicate the x-axis center point of the bounding box, the y-axis center point of the bounding box, the x-axis velocity, and the y-axis velocity, respectively. Likewise, $Z_k$ and $J_{k}$ denote a set of observations (detection responses) and the number of them at time $k$, respectively. An observation $z_k$ is composed of $(c_x,c_y)$, where $c_x$ and $c_y$ indicate the x-axis and the y-axis center of the detection bounding box, respectively. 
Equation~\eqref{eq3} and~\eqref{eq4} describe the basic notations of state and observation.
\begin{align}
&x^i_k = \{c_{x,k},c_{y,k},v_{x,k},v_{y,k}\}^T, \label{eq3}\\
&z^j_k = \{c_{x,k},c_{y,k}\}^T. \label{eq4}
\end{align}
The tracking process of the GM-PHD filter is composed of four steps: \textit{Initialization}, \textit{Prediction}, \textit{Update}, and \textit{Pruning} as follows.

\textit{\textbf{Initialization}}:
\begin{align}
    \sum _{ i=1 }^{ { I }_{ 0 } }{ { w }_{ 0 }^{ i }\mathcal{N}(x;{ m }_{ 0 }^{ i },{ P }_{ 0 }^{ i }) }, \label{eq5}
\end{align}
where the GMM is initialized by the initial observations from the detection responses. Besides, when an observation fails to find the association pair, i.e., updating object state, the observation initializes a new Gaussian model (a new state). 
Gaussian probability function $\mathcal{N}$ represents tracking objects with weight $w$, mean vector $m$, object state vector $x$, and covariance matrix $P$. At this step, we set the initial velocities of mean vector to zeros. Each weight is set to the normalized confidence value of the corresponding detection response.

\textit{\textbf{Prediction}}:
\begin{align}
    &\sum _{ i=1 }^{ { I }_{ k-1 } }{ { w }_{ k-1 }^{ i }\mathcal{N}(x;{ m }_{ k-1 }^{ i },{ P }_{ k-1 }^{ i }) }, \label{eq6}\\
    &{ m }_{ k|k-1 }^{ i }={ F }{ m }_{ k-1 }^{ i },\label{eq7}\\
    &{ P }_{ k|k-1 }^{ i }={ Q }+{ F }{ P }_{ k-1 }^{ i }{ { (F }) }^{ T },\label{eq8}
\end{align}
where we assume that the GMM representing the objects' states was initialized or active at the previous frame $k-1$ in~\eqref{eq6}. 
In~\eqref{eq7} and~\eqref{eq8}, $F$ is the state transition matrix and $Q$ is the process noise covariance matrix. $F$ and $Q$ are constants in our tracker.
Then, we can predict the state at time $k$ using the Kalman filtering. In~\eqref{eq7}, $m_{k|k-1}^i$ is derived by using the velocity of $m_{k-1}^i$. Covariance $P_{k|k-1}^i$ is also predicted by the Kalman filtering method in~\eqref{eq8}.   

\textit{\textbf{Update}}:
\begin{align}
 &\sum _{ i=1 }^{ { I }_{ k|k-1 } }{ { w }_{ k }^{ i }(z)\mathcal{N}(x;{ m }_{ k|k }^{ i },{ P }_{ k|k }^{ i }) }, \label{eq9}\\
 &{q}_{k}^{i}(z)=\mathcal{N}(z;{H}{m}_{k|k-1}^{i},{R}+{H}{P}_{k|k-1}^{i}{({H})}^{T}),\label{eq10}\\
 &{w}_{k}^{i}(z)=\frac {{w}_{k|k-1}^{i}{q}_{k}^{i}(z)}{\sum _{l=1}^{{I}_{k|k-1}}{{w}_{k|k-1}^{l}{q}_{k}^{l}(z)}  }, \label{eq11}\\
 &{ m }_{ k|k }^{ i }(z)={ m }_{ k|k-1 }^{ i }+{ K }_{ k }^{ i }{ (z-{ H }{ m }_{ k|k-1 }^{ i }) },\label{eq12}\\
 &P_{k|k}^i=[I-{K_k}^{i}{H}]{P_{k|k-1}^i},\label{eq13}\\
 &K_k^i=P_{k|k-1}^i(H)^T({H}P_{k|k-1}^i(H)^T+R)^{-1},\label{eq14}
\end{align}
where the goal of update step is deriving~\eqref{eq9}. First, we should find an optimal observation $z_k$ at time $k$ to update a Gaussian model. The optimal $z$ makes $q_k$ be the maximum in~\eqref{eq10}. $R$ denotes the observation noise covariance. $H$ denotes the observation matrix to transit a state vector to an observation vector. Both $R$ and $H$ are constants in our application.
In the perspective of application, the update step involves data association. Updating the Gaussian state models follows finding the optimal observations updating the states through the data association. After finding the optimal $z$, the GMM is updated to~\eqref{eq9} through~\eqref{eq10},~\eqref{eq11}, and~\eqref{eq14},~\eqref{eq13},~\eqref{eq12}.

\textit{\textbf{Pruning}}:
\begin{align}
&\tilde{X_k} = \{m_k^i:w_k^i\ge\theta_w,i=1,...,I_k\},\label{eq15}\\
&\tilde{W_k} = \{w_k^i:m_k^i \in \tilde{X_k}, i=1,...,I_k\},\label{eq16}\\
&\tilde{W_k} = \{\tilde{w_{k,1}},...,\tilde{w}_{k,{\tilde{I}_k}}\}, \tilde{I_k} = |\tilde{W_k}|,\label{eq17}\\
&{w}_{k}^{i}=\frac {\tilde{w}_{k}^{i}}{\sum_{l=1}^{\tilde{I_k}}{\tilde{w}_{k}^{l}}},\label{eq18}\\
&{X_k} = \tilde{X_k},\label{eq19}
\end{align}
where the states with the weight under threshold $\theta_w$ are pruned as in~\eqref{eq15}. We experimentally set $\theta_w$ to $0.1$. Then the weights of the surviving states are normalized as shown in~\eqref{eq18}. The pruning step handles the false positive tracks by the false positive detections.

The GMPHD filter~\cite{gmphd} is specialized in handling false positives e.g., clutters and noise.
However, tracking systems have the different problems, depending on their domains as shown in Figure~\ref{fig:domains}, where input and output indicate detection results (observations) and tracking results (states), respectively.
As presented in ~\cite{Vo_thesis} and Figure~\ref{fig:domains}-(a), at radar/sonar systems, the senors receive uncountable detection responses with a lot of clutters but objects are rarely missed. On the other hand, as shown in Figure~\ref{fig:domains}-(b), the video data based detectors observe less clutters and miss more objects than the radar/sonar senors do.
The conventional GMPHD filter is effective to handle the clutters (false positive) but missed detctions cause the new tracking problems in video data system (false negative).
Thus, we propose the GMPHD filtering based tracker with a hierarchical data association strategy. 

\subsection{Hierarchical Data Association}
\label{hda}
Video-based tracking systems have inherent problems as shown in Figure~\ref{fig:domains}-(b). 
Generally, when objects are not detected, the objects' IDs are frequently changed and the tracks are fragmented if only detection-to-track association is employed. 
To prevent these problems by missing objects, we take advantage of a hierarchical data association (HDA) strategy which has been widely used in many online multi-object tracking methods~\cite{bae1,bae2,prev1,prev2,eamtt}. Thus, in this paper, we propose a simple HDA scheme with just two stages. 
The proposed HDA includes detection-to-track (D2T) and track-to-track (T2T) associations.
We implement the both association methods with the GMPHD filtering process as given in~\ref{subsec:gmphd}.
Also, we derive a cost function from~\eqref{eq11} of the GMPHD filtering process as follows:
\begin{align}
    Cost(x^i_{k|k-1},z_k^j)=-\ln w^i_k(z_k^j),\label{eq20}
\end{align}
where $w^i_k$ indicate the weight value, assuming that observation $z_k^j$ updates state $x^i_{k|k-1}$. We use $-\ln w^i_k(z_k^j)$ as a cost between $x^i_{k|k-1}$ and $z_k^j$.
Then, cost matrix $C$ can be built by every pair between state set $X_{k|k-1}$ and observation set $Z_{k}$ as follows:
\begin{align}
    C[i,j] =  Cost(X_{k|k-1}[i],Z_{k}[j]).\label{eq21}
\end{align}
When the cost matrix C is built, the Hungarian algorithm is used to solve it. 
Then, the optimal pairs between observations and states are found, and consequently state $x_{k|k-1}$ is updated to $x_k$ in D2T and T2T associations.
In~\ref{DA:D2TA}and~\ref{DA:T2TA}, we introduce the definition of observations and states in each association stage with more detail usage of the cost function.

\subsubsection{Detection-to-Track Association (D2TA, Stage 1)}
\label{DA:D2TA}
In D2TA, observation set $Z_k$ is filled with detection responses at time $k$. We assume that state set $X_{k-1}$ already exists from time $k-1$, and then $X_{k|k-1}$ is predicted by using the Kalman filtering as shown in~\eqref{eq6}-\eqref{eq8}. Thus, the cost matrix $C_{D2T}$ is easily calculated with these sets $X_{k|k-1}$ and $Z_k$.
\subsubsection{Track-to-Track Association (T2TA, Stage 2)}
\label{DA:T2TA}
In T2TA, a simple temporal analysis of tracklet is conduced. A tracklet means a fragment of the track, and becomes a calculation unit.
Before T2TA, all tracklets are categorized into two types, according to success or failure of tracking at the present time $k$ as follows:
\begin{align}
&T^{lost}_k \cup T^{live}_k = T^{all}_k,\label{eq22}\\
&T^{lost}_k \cap T^{live}_k = \phi,\label{eq23}\\
&T^{lost}_k = \{\tau^{lost}_{1,k},...,\tau^{lost}_{i,k}\},\label{eq24}\\
&\tau^{lost}_{i,k} = \{a^i_s,..,a^i_{t}\},\quad\quad\,\,\,\, 0\le s<t<k,\label{eq25}\\
&T^{live}_k = \{\tau^{live}_{1,k},...,\tau^{live}_{j,k}\},\label{eq26}\\
&\tau^{live}_{j,k} = \{a^j_s,..,a^j_{t}\},\quad\quad\,\,\,\, 0\le s<t,\,t=k,\label{eq27}
\end{align}
where \textit{``live"} indicates that tracking succeeds at time $k$. \textit{``lost"} indicates that tracking fails at time $k$. Then, for the T2TA, observation set $Z_k$ is filled with the first (oldest) elements $a^j_s$s of \textit{``live"} tracklets. However, the state set $X_{k|k-1}$ is not filled with the last (most recent) elements $a^i_t$s of \textit{``lost"} tracklets. One prediction step is needed as follows:
\begin{align}
&a^i_s = \{c_{x,s},c_{y,s},v_{x,s},v_{y,s}\}^T,\label{eq28}\\
&a^i_t = \{c_{x,t},c_{y,t},v_{x,t},v_{y,t}\}^T,\label{eq29}\\
&x^i_{t} = \{c_{x,t},c_{y,t},\frac {c_{x,t}-c_{x,s}}{t-s},\frac{c_{y,t}-c_{y,s}}{t-s}\}^T,\label{eq30}\\
&x^i_{k|k-1} = F^{T2T}x^i_{t},\label{eq31}\\
&F^{T2T} = \begin{pmatrix} 1 & 0 & d_f & 0  \\ 0 & 1 & 0 & d_f  \\ 0 & 0 & 1 & 0  \\ 0 & 0 & 0 & 1  \end{pmatrix},\label{eq32}\\
&d_{f}(i,j) = \text{frame difference between } a^i_t \text{ and } a^j_s.\label{eq33}
\end{align}
In (30) $\frac {c_{x,t}-c_{x,s}}{t-s}$ and $\frac{c_{y,t}-c_{y,s}}{t-s}$ are the averaged velocities in terms of x-axis and y-axis, respectively. The velocities are calculated by subtracting the center position of the first object state $a^i_s$ from that of the last state $a^i_t$, and dividing it by the frame difference $t-s$ which is equivalent to the length of \textit{``lost"} tracklet $\tau^{lost}_{i,k}$. D2TA has the identical time interval \textit{``1"} between states and observations in transition matrix $F$, whereas in T2TA , each cost of matrix $C_{T2T}$ has different time interval (frame difference) between states and observations. Variable $d_f$ depends on which state of \textit{``lost"} tracklet and observation of \textit{``live"} tracklet are paired. \eqref{eq31} means the prediction process of state with linear motion analysis. Finally, the cost matrix $C_{T2T}$ is filled by~\eqref{eq31} and the oldest element $a^j_s$ of live tracklet $\tau_{j,k}^{live}$.

The pseudo-code in Algorithm~\ref{algo:proposed} includes the procedures presented in this section. \textit{Initialization}, \textit{Prediction}, Cost-minimization, \textit{Update}, and \textit{Pruning} in D2TA correspond to each of line 5-10, 13-15, 16-21, 22-24, and 25-27 in Algorithm~\ref{algo:proposed}. Tracklet-categorization, Cost-minimization, \textit{Update} in T2TA correspond to line 36-44, 49-54, and 55-66 in Algorithm~\ref{algo:proposed}, respectively.

\begin{algorithm}[h]
\footnotesize 
\caption{Proposed Online MOT Algorithm}
\label{algo:proposed}

\Comment{$k$ : the current frame number}

\Comment{$X_{k-1}$ : a set of states at time $k-1$}

\Comment{$Z_k$ : a set of observations at time $k$}

\Comment{$\sigma_{m}$ : threshold for track merging}

\Comment{$\tau_{T2T}$ : the minimum track length for T2TA}

\Comment{$\theta_{T2T}$ : the maximum frame interval for T2TA}

\Comment{$T^{live}$ : a \{key:id,value:tracklet\} set of live tracklets}

\Comment{$T^{lost}$ : a \{key:id,value:tracklet\} set of lost tracklets}
\begin{algorithmic}[1]
\,\,
\Procedure{GMPHD\_OGM}{$k$,$X_{k-1}$,$Z_k$,$\sigma_{m}$,$\tau_{T2T}$,$\theta_{T2T}$,$T^{live}$,$T^{lost}$}
\State $l=|X_{k-1}|$;\hfill \textit{// the number of states}
\State $m=|Z_{k}|$;\hfill \textit{// the number of observations}
\State $G_{k-1},G_k$; \hfill \textit{// a set of occlusion groups at time k-1 and k.}
\,\,\If {$k=1$ or $l=0$}
        \State Initialize states $X'_k$ with $Z_k$;
        \State $G_{k-1} = G_k$;
        \State $X_k = MERGE(X'_k,\sigma_{m},G_k)$;
        \State \Return $X_k$;
    \EndIf

\,\,\LeftComment{1. Detection-to-Track Association (D2TA)}
\,\,\State $C_{D2T}[1 \ldots {l}][1 \ldots {m}]$; \hfill \textit{// for cost matrix}
\State $P_{D2T}[1 \ldots {l}]$; \hfill \textit{// for pairing observations' indices}
\,\,\indentComment{1}{/*predict states $X_{k-1}$ to be $X_{k|k-1}$*/}

\For{$i = 1$ to ${l}$} 
    \State $X_{k|k-1}[i] = PREDICT(X_{k-1}[i])$;
\EndFor
\indentComment{1}{/*calculate the GMPHD filter cost matrix $C_{D2T}$*/}
\For{$i = 1$ to ${l}$} 
    \For{$j = 1$ to ${m}$}
    \State $C_{D2T}[i][j]=COST_{D2T}(X_{k|k-1}[i],Z_k[j])$;
    \EndFor
\EndFor

\,\,\indentComment{1}{/*find min-cost pairs by the Hungarian method*/}
\State $P_{D2T}=HungrianMethod(C_{D2T})$;

\,\,\indentComment{1}{/*update and birth states*/}
\indentComment{1}{/*update $X_{k|k-1}$ with the min-costly observations*/}
\For{$i = 1$ to ${l}$} 
    \State $X'_k[i] = UPDATE(X_{k|k-1}[i],Z_k[P_{D2T}[i]])$;
\EndFor
\indentComment{1}{/*prune $X_{k|k-1}$ with the weight under 0.1*/}
\For{$i = 1$ to ${l}$} 
    \State $X'_k[i] = PRUNE(X_{k|k-1}[i])$;
\EndFor
\For{$j = 1$ to ${m}$}
    \If {$Z_k[j]$ is not assigned to update any state}
        \State Initialize newly birth state $x$ with $Z_k[j]$;
        \State $X'_k$ = $X'_k\cup\{x\}$;
    \EndIf
\EndFor
\,\,\LeftComment{2. Merge States and Find Occlusion Groups}
\State $G_{k-1} = G_k$;
\State $X_k = MERGE(X'_k,\sigma_{m},G_k)$;
\,\,\indentComment{1}{/*manage tracklet pool after D2TA and MERGE*/}
\For{$i = 1$ to $|X_k|$}
    \If{$X_k[i]$ is active}
        \State update $T^{live}[X_k[i].id]$ with $X_k[i]$;
        \State delete $T^{lost}[X_k[i].id]$;
    \Else
        \State update $T^{lost}[X_k[i].id]$ with $X_k[i]$;
        \State delete $T^{live}[X_k[i].id]$;
    \EndIf
\EndFor
\algstore{myTracker}
\end{algorithmic}
\end{algorithm}

\begin{algorithm}[h]
\begin{algorithmic}[1]
\footnotesize
\algrestore{myTracker}
\,\,\LeftComment{3. Track-to-Track Association (T2TA)}
\,\,\State $t_{1}=|T^{lost}|$;\hfill \textit{// the number of lost tracklets}
\State $t_{2}=|T^{live}|$;\hfill \textit{// the number of live tracklets}
\State $C_{T2T}[1 \ldots {t_{1}}][1 \ldots {t_{2}}]$; \hfill \textit{// for cost matrix}
\State $P_{T2T}[1 \ldots {t_{1}}]$; \hfill \textit{// for pairing observations' indices}
\,\,\indentComment{1}{/*calculate the GMPHD filter cost matrix $C_{T2T}$*/}
\For{$i = 1$ to $t_{1}$}
    \For{$j = 1$ to $t_{2}$}
    
    \State $C_{T2T}[i][j]=COST_{T2T}(T^{lost}[i],T^{live}[j],\tau_{T2T},\theta_{T2T})$;
    \EndFor
\EndFor
\,\,\indentComment{1}{/*find min-cost pairs by the Hungarian method*/}
\State $P_{T2T}=HungrianMethod(C_{T2T})$;
\indentComment{1}{/*update tracklets and manage tracklet pool after T2TA*/}
\For{$i = 1$ to ${l}$} 
    \State $X'_k[i] = UPDATE(X_{k|k-1}[i],Z_k[P_{D2T}])$;
\EndFor
\For{$i = 1$ to $|X_k|$}
    \If{$X_k[i]$ is active}
        \State update $T^{live}[X_k[i].id]$ with $X_k[i]$;
        \State delete $T^{lost}[X_k[i].id]$;
    \Else
        \State update $T^{lost}[X_k[i].id]$ with $X_k[i]$;
        \State delete $T^{live}[X_k[i].id]$;
    \EndIf
\EndFor
\,\,\LeftComment{4. Occlusion Group Energy Minimization (OGEM)}
\If{$k>1$ and $|G_{K-1}|>0$}
    \State $OGEM(k,G_{k-1},X_{k})$;
    \,\,\indentComment{1}{/*manage tracklet pool after OGEM*/}
\For{$i = 1$ to $|X_k|$}
    \If{$X_k[i]$ is active}
        \State update $T^{live}[X_k[i].id]$ with $X_k[i]$;
        \State delete $T^{lost}[X_k[i].id]$;
    \Else
        \State update $T^{lost}[X_k[i].id]$ with $X_k[i]$;
        \State delete $T^{live}[X_k[i].id]$;
    \EndIf
\EndFor
\EndIf
\State \Return $X_k$;\hfill \textit{// return final states $X_{k}$}
\EndProcedure

\end{algorithmic}
\end{algorithm}

\section{Occlusion Group Management Scheme}
\label{Contribution}
\begin{figure}[t]
\centering
\includegraphics[width=8.5cm]{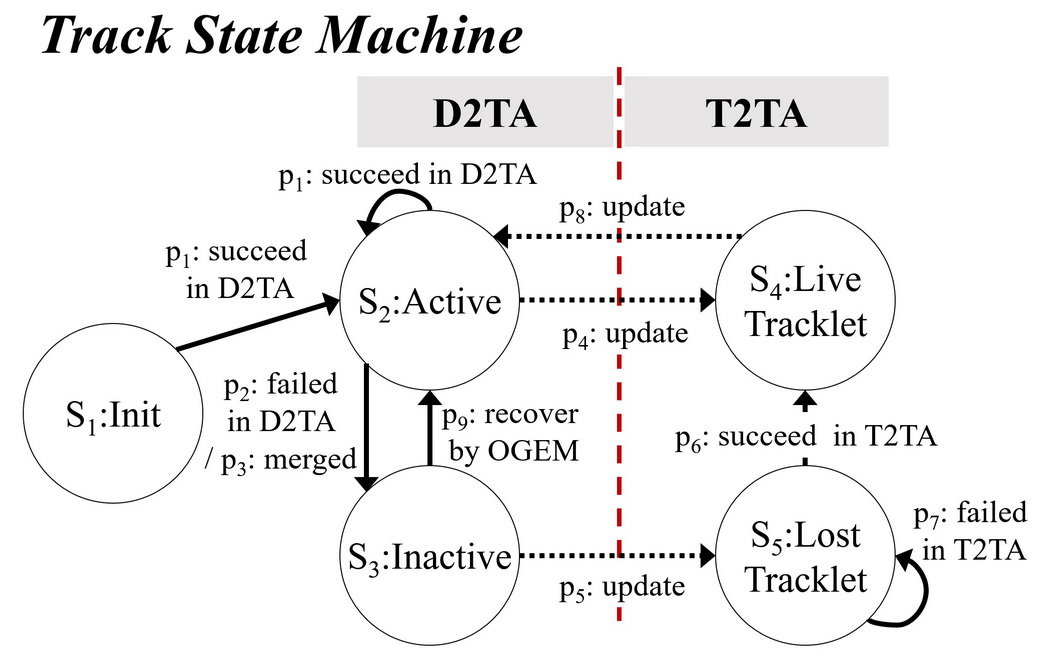}
\caption{In the proposed tracking framework, an object with state $S_1:Init$ is transited within the defined states $\{S_2,S_3,S_4,S_5\}$ by the state-transition functions $\{p_1,p_2,p_3,p_4,p_5,p_6,p_7,p_8,p_9\}$.}
\label{fig:state_machine}
\end{figure}

In Section~\ref{Framework}, we addressed that the proposed online multi-object tracking framework is based on the GMPHD filtering theory with the two-stage hierarchical data association. However, the tracking results from that framework still give uncertainty to us, even if we effectively extend the conventional GMPHD filter to be suitable for video-based tracking system. Thus, to handle it, we define two types of tracking problems and provide a solution. One is an intrinsic occlusion and the other is an extrinsic occlusion. The intrinsic occlusion is defined when the number of detection responses on one object is more than one. Generally, it makes the object ID switched and false positive tracks. The extrinsic occlusion is defined when the number of detection responses on objects, occluded each other, is less or more than the number of the occluded objects. That can cause false negative and positive tracks, respectively. 
Figure~\ref{fig:merge} and \ref{fig:qual_results} show the defined tracking issues well. 
The false positive detections made by intrinsic and extrinsic occlusions inevitably generate false tracks as shown in the second row Figure~\ref{fig:qual_results} (D2TA), if appropriate techniques do not handle it.  
To resolve the two types of problems, we design a new occlusion group management (OGM) scheme.
OGM consists of \textit{``Track Merging"} and \textit{``Occlusion Group Energy Minimization (OGEM)"} routines which execute just after D2TA and T2TA, respectively. 
Figure~\ref{fig:framework} briefly shows the tracking pipeline with those two components of OGM.
Consequently, our occlusion group management technique not only decreases false positive tracking results 
but also prevents occluded tracks from false ``track merging".
The effectiveness of the proposed OGM method is discussed in Section~\ref{Experiments} in more detail.

\begin{algorithm}[t]
  \footnotesize
   \caption{Track Merging using the SIOA Metric}
   \label{algo:merge}
      \Comment{$X_k$ : a set of states at time $k$}
      
      \Comment{$\sigma_{m}$ : threshold for merging}
      
      \Comment{$G_k$ : a set \{key:id,value:states\} of occlusion groups at time k}
    \begin{algorithmic}[1]
      \Function{Merge}{$X_k$,$\sigma_{m}$,$G_k$}
        \State ${l} = |X_k|$; \hfill \textit{// $l$ : the number of states $X_k$}
        
        \State Let $M[1 \ldots {l}][1 \ldots {l}]$ be the array set to all \textit{false};
        
        \,\,\LeftComment{measure occlusion ratio between all states \\ by using the SIOA metric}
        \For{$i = 1$ to ${l}$} 
            \For{$j = i+1$ to ${l}$}
                \State $r_{occ}=SIOA_{X_k[i], X_k[j]}$;\hfill \textit{// SIOA occlusion ratio.}
                \If{$r_{occ} > \sigma_{m}$}
                    \State $M[i][j]=true$;\hfill \textit{// check to be merged}
                    \State $M[j][i]=true$;\hfill \textit{// double check}
                \ElsIf{$r_{occ} \le \sigma_{m}$ and $r_{occ} > 0$}
                    \State $id_i=X_k[i].id,id_j=X_k[j].id$;
                    \If{$id_i<id_j$}
                        \State $G_k[id_i] = G_k[id_i] \cup \{X_k[i],X_k[j]\}$;
                    \Else
                        \State $G_k[id_j] = G_k[id_j] \cup \{X_k[i],X_k[j]\}$;
                    \EndIf
                \EndIf
            \EndFor
        \EndFor
        \,\,\LeftComment{merge the states where $SIOA\,\,value > \sigma_{m}$}
        \For{$i = 1$ to ${l}$} 
            \For{$j = i+1$ to ${l}$}
                \If {$M[i][j]=true$}
                    \If {$X_k[i].id < X_k[j].id$}
                        \State $X_k[i]={0.9}*{X_k[i]}+{0.1}*{X_k[j]}$;
                        \State Deactivate state $X_k[j]$;
                    \Else
                        \State $X_k[j]={0.9}*{X_k[j]}+{0.1}*{X_k[i]}$;
                        \State Deactivate state $X_k[i]$;
                    \EndIf
                \EndIf
            \EndFor
        \EndFor
        \State \Return $X_k$;
       \EndFunction

\end{algorithmic}
\end{algorithm}

\subsection{Track Merging}
\label{merging}
Merging the neighboring objects' states with the distances under a threshold is proposed in~\cite{gmphd} already.
However, it can only reflect point-to-point distance without considering regional information e.g., overlapping ratio between visual objects (bounding boxes). To measure the overlapping ratio, 
the intersection-over-union (IOU) metric has been widely used which was originally designed to measure mAP in object detection research fields~\cite{voc,imagenet}.
However, the IOU metric is nice to refine the detection bounding boxes but not adjustable to measure overlapping ratio for merging the objects.
Figure~\ref{fig:merge} explains that reason by a case study.
The case study mainly assumes that the number of detection responses (observations) is larger than the number of real objects. When the observations most likely include false positive detections, the object states by those observations also most likely become the false positive states.
So, to handle and consider the characteristic of those observations with the false positive errors, we propose a new metric named as sum-of-intersection-over-area (SIOA).
The IOU and SIOA metrics are formulated as follows:
\begin{align}
\label{eq34:iou}
     & IOU_{AB} = {\frac{area(A) \cap area(B)}{area(A){\cup}area(B)}},\\
     & \nonumber\\
\label{eq35:sioa}
     & SIOA_{AB} = \\
     & 0.5*({\frac{area(A) \cap area(B)}{area(A))}}+{\frac{area(A) \cap area(B)}{area(B))}})\nonumber,
\end{align}
where A and B indicate two different objects. $area$ represent a bounding box $(x,y,width,height)$.
Algorithm \ref{algo:merge} describes the proposed track merging method.
Track merging with the SIOA metric follows after the D2T association as presented in Subsection \ref{DA:D2TA} and Figure~\ref{fig:framework}.

\begin{algorithm}[t]
\footnotesize
   \caption{Occlusion Group Energy Minimization}
   \label{algo:ogem}
        \Comment{$k$ : the current frame number}
   
        \Comment{$G_{k-1}$ : a set of occlusion groups at time $k-1$}
        
        \Comment{$X_k$ : a set of states at time $k$}
      
    \begin{algorithmic}[1]
      \Function{OGEM}{$k$,$G_{k-1}$,$X_k$}
        \State ${l} = |G_{k-1}|$; \hfill \textit{// $l$ : the number of the groups $G_{k-1}$.}
      
        \State ${n} = |X_k|$; \hfill \textit{// $n$ : the number of the states $X_k$.}
        
        \,\,\indentComment{1}{/*build the GMMs for all occlusion groups at time k-1*/}
        \,\,\indentComment{1}{/*a GMM is used for the defined energy function in~\eqref{eq36}*/}
        \For{$i = 1$ to ${l}$} 
        
            \State  $p_i = ^{|G_{k-1}[i]|}P_2$ \hfill \textit{//the number of topological vectors.}
            \State  $GMM[1 \ldots {p_i}];$ \hfill \textit{//the Gaussian mixture for a group.}

            \For{$j = 1$ to ${p_i}$}  \hfill \textit{//iterate topologies in a group.}
                \State Initialize a Gaussian mixture $GMM[j]$ with
                \State the mean vectors $m$ having topological info and 
                \State the covariance matrix $R$ as defined in~\eqref{eq36}
            \EndFor
            \State $E_{min} = DBL\_MAX$;     \hfill \textit{//variable for the min-energy.}
            \State $h_{min} = 0$;   \hfill \textit{//index to the optimal hypothesis.}
            \For{$h = 1$ to $|H|$} \hfill \textit{//iterate topological hypotheses.}
                \If{$E(h)<E_{min}$} \hfill \textit{//find the optimal hypothesis.}
                    \State $h_{min}=h;$
                    \State $E_{min}=E(h);$
                \EndIf
            \EndFor
            \State Update $G_{k-1}[i]$ with $h_{min}$;
        \For{$g$ in $G_{k-1}[i]$} \hfill \textit{//iterate group $G_{k-1}[i]$.}
            \State Find the state $x$ with $g.id$ in $X_k$;
            \If{$x$ is in $X_k$}
                \State $X_k[g.id] = g$;
            \Else
                \State $X_k = X_k \cup \{g\}$;
            \EndIf
            \EndFor
        \EndFor
        \State \Return $X_k$;
       \EndFunction

\end{algorithmic}
\end{algorithm}

\subsection{Occlusion Group Energy Minimization}
\label{subsec:ogem}
\begin{figure}
\centering
\includegraphics[width=8.5cm]{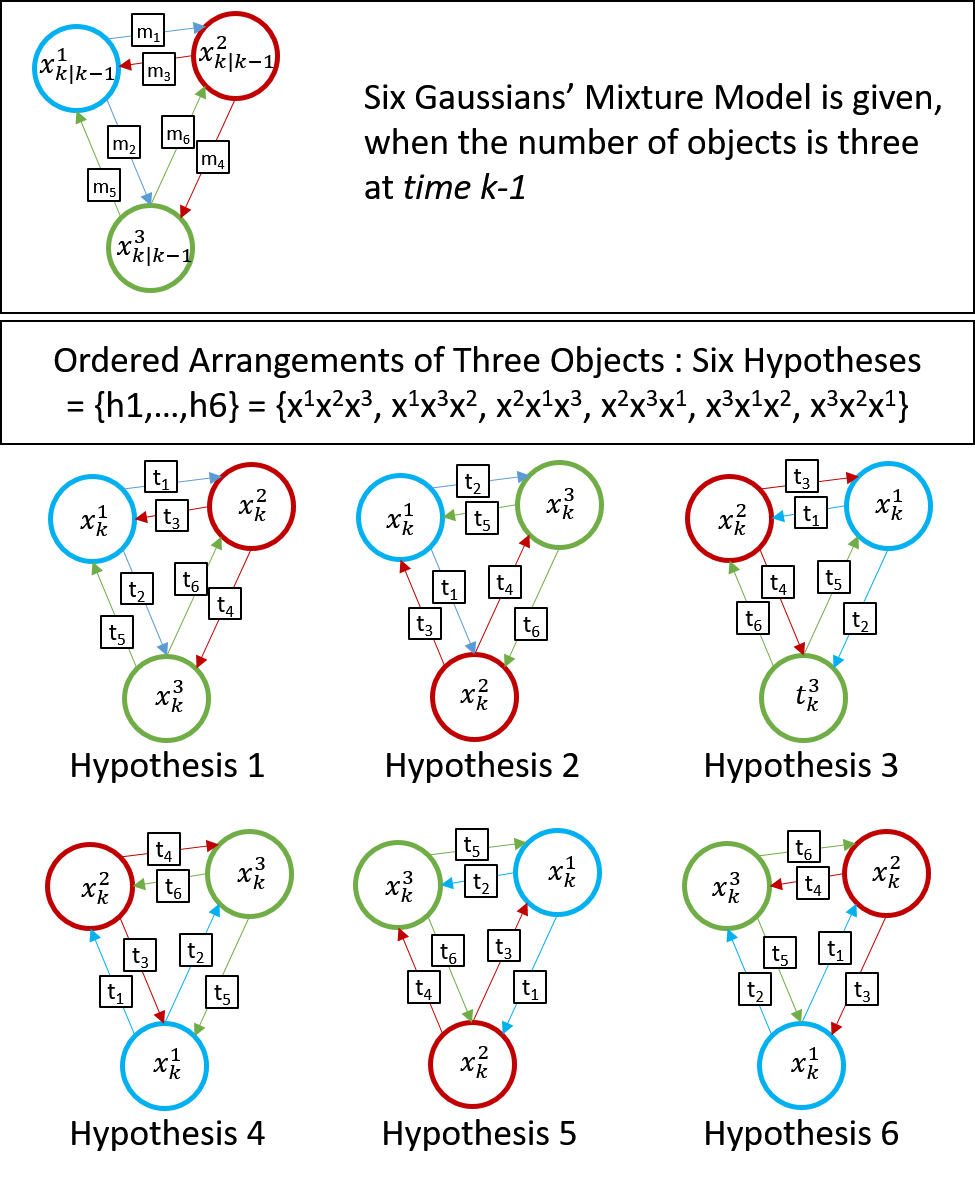}
\caption{Illustration of the proposed occlusion group energy minimization represented by the Gaussian mixture model. Six hypotheses exist in the case of three occluded objects.}
\label{fig:gem}
\end{figure}

The occlusion group energy minimization method is devised to prevent the true objects which are occluded to others from false merging. In other words, track merging after detection-to-track association (D2TA) may merge occluded objects with correct number of observations into the states with less number of real objects. That can cause tracking errors such as false negative and fragmented tracks. 

Thus, we propose a new energy minimization model to prevent false merging, named as \textit{``Occlusion Group Energy Minimization (OGEM)"}. Each group of occluded objects has an energy function represented by a Gaussian mixture model (GMM) as follows:
\begin{align}
E(h) = -\ln{\sum \mathcal{N}(t|m,R)},\label{eq36}
\end{align}
where $h$, $t$, $m$, and $R$ indicate hypothesis, topological vector, mean vector, and Gaussian covariance marix (noise), respectively. A Gaussian probability function $\mathcal{N}$, i.e., component of the GMM, indicates a topological position vector between two objects in an occlusion group, which is given at time $k-1$. The Gaussian function has a mean vector $m$ which denotes the topological position, i.e., relative position between the predicted center positions at time $k$ of the objects in the group. Those objects are denoted by $x_{k|k-1}$ and the notation ${k}|{k-1}$ indicates the prediction at time $k$ from position and velocity at time $k-1$. 
If there are three occluded objects in a group, six hypotheses exists as shown in Figure~\ref{fig:gem}. One hypothesis is a set of six topological vectors $\{t_1,t_2,t_3,t_4,t_5,t_6\}$. For example, $m_1$ is calculated by ${x^2_{k|k-1}}-{x^1_{k|k-1}}$ and $t_1$ is calculated by ${x^2_k}-{x^1_k}$. In the case that an object state $x^d_k$ becomes inactive by false merging in occlusion, we build a new hypothesis using a $x^d_{k|k-1}$as a dummy. Then the dummy added hypothesis recovers the false merged object.
If there are $n$ occluded objects in a group, $n(n-1)$ hypotheses exists with the condition $1<n<4$.
Then with these topological models, we can find an optimal one among all hypotheses making the group cost minimal.

Whereas track merging runs after the D2TA, the proposed occlusion group energy minimization follows Track-to-Track Association (T2TA) as described in Figure~\ref{fig:framework}.
Figure~\ref{fig:qual_results} includes some examples to explain that the proposed group energy minimization complements track merging step. The tracking stage at frame 42 explains it.

In summary, both \textit{``Track Merging"} and \textit{``Oclcusion Group Energy Minimization"} procedures assume occlusion situations, and the GMPHD filtering is adopted as the main framework. The pseudo-code examples of proposed occlusion group management scheme are described in Algorithm~\ref{algo:merge} and~\ref{algo:ogem}. Also, both methods correspond to line 8, 35 and 67-78 in Algorithm~\ref{algo:proposed} which represents the whole tracking framework.
From now on we use GMPHD-OGM as the abbreviation for the proposed algorithm, online multi-object tracking with the GMPHD filter and occlusion group management.

\section{Experiments}
\label{Experiments}

In this section, we present development environment including parameter settings, and also discuss evaluation results of the GMPHD-OGM tracker which include an ablation study with baselines and comparisons to state-of-the-art methods.
The GMPHD-OGM tracker is implemented by Visual C++ with OpenCV3.4.1 and boost1.61.0 libraries, and without any GPU-accelerated libraries such as CUDA.
All experiments are conducted on Windows 10 with Intel i7-7700K CPU @ 4.20GHz and DDR4 32.0GB RAM.

\subsection{Parameter Setting}
\label{subsec:params}
Our proposed tracking framework involves several parameter settings.
Parameter $\sigma_{m}$ indicates the threshold for ``Track Merging" which is set to $0.5$ in terms of the SIOA metric.
$0.5$ is set not only empirically set but by considering the occlusion cases between size-variant bounding boxes as shown in Case 4 and 5 of Figure~\ref{fig:merge}.
$\tau_{T2T}$ and $\theta_{T2T}$ are related to track-to-track association (T2TA) of the hierarchical data association, whose parameters are selected adaptively, scene-by-scene. 
The optimal values of $\tau_{T2T}$ and $\theta_{T2T}$ are gained from the ablation study presented in Figure~\ref{fig:ablation}. We use the optimal parameter settings from training to test sequences. 
These three parameters are summarized in Table~\ref{parameters}. 

The GMPHD filtering process has a set of static parameters.
The matrices $F$, $Q$, $P$, $R$, and $H$ are used in \textit{Prediction Step} and  \textit{Update Step}.
Also, $\theta_w$ is used in \textit{Pruning Step}.
Experimentally, we set the parameters for the GMPHD filter's tracking process as follows:

\begin{align*}
    &F = {\footnotesize\begin{pmatrix} 1 & 0 & 1 & 0  \\ 0 & 1 & 0 & 1  \\ 0 & 0 & 1 & 0  \\ 0 & 0 & 0 & 1  \end{pmatrix}},
    Q = {\footnotesize\dfrac{1}{2}}
    {\footnotesize\begin{pmatrix} 5^{ 2 } & 0 & 0 & 0  \\ 0 & 10^{ 2 } & 0 & 0  \\ 0 & 0 & 5^{ 2 } & 0  \\ 0 & 0 & 0 & 10^{ 2 } \end{pmatrix}},  \\
    &P = {\footnotesize\begin{pmatrix} 5^{ 2 } & 0 & 0 & 0  \\ 0 & 10^{ 2 } & 0 & 0  \\ 0 & 0 & 5^{ 2 } & 0  \\0 & 0 & 0 & 10^{ 2 }  \end{pmatrix}},
    R = {\footnotesize\begin{pmatrix} 5^{ 2 } & 0  \\ 0 & 10^{ 2 } \end{pmatrix}}, \\
    &H ={\footnotesize\begin{pmatrix}1 & 0 & 0 & 0  \\ 0 & 1 & 0 & 0 \end{pmatrix}}, \theta_w = 0.1,
\end{align*}

\begin{table}[b]
\caption{Parameter Settings for ``Track Merging" and track-to-track association (T2TA).}
\label{parameters}
\setlength{\tabcolsep}{3pt}
\begin{tabular}{|p{25pt}|p{150pt}|p{30pt}|}
\hline
\textbf{Symbol} & 
\textbf{Description} & 
\textbf{Value} \\
\hline
$\sigma_{m} $& 
threshold for track merging. &
$0.5$ \\
\hline
$\tau_{T2T}$& 
the minimum track length for T2TA & 
${1, 2, 3}$ \\
\hline
$\theta_{T2T}$&
the maximum frame interval for T2TA & 
$5$ to $100$ \\
\hline
\end{tabular}
\label{tab1}
\end{table}

\subsection{Evaluation Results}

In this section, we evaluated the propose method with state-of-the-art online~\cite{bae2,ham,fu1,mdp,tbss,scea,eamtt,rmot,prev1,gscr,bae1,motdt,dman,mtdf,amadm,gmphdn1tr,gmphdkcf,gmphd2012} and offline~\cite{mhtdam,cnntcm,siameseCNN,nomt,elp,jpdam,moticon,segtrack,cem,smot,dpnms,mhtlstm,ehaf,fwt,jcc,tlmht,edmt,iou} MOT methods in terms of the CLEAR-MOT metrics~\cite{clearmot}. The CLEAR-MOT metrics gracefully measure multi-object tracking performance from the detailed perspectives such as multi-object tracking accuracy (MOTA),  multi-object tracking precision (MOTP), mostly tracked objects (MT), mostly lost objects (ML), the total number of flase positives (FP), the total number of false negatives (missed tracks, FN), the total number of identity switches (IDS), the total number of times that a trajectory is fragmented (Frag), and processing speed (frames per second, FPS). Among these metrics, MOTA is normally proposed as the key metric, because it considers three error sources including FP, FN, and IDS, comprehensively.
The evaluation results contain not only the tracking results on the MOT15 and MOT17 test datasets but also an ablation study on the MOT15 training dataset. 

\begin{figure*}[h]
\centering
\includegraphics[height=20cm]{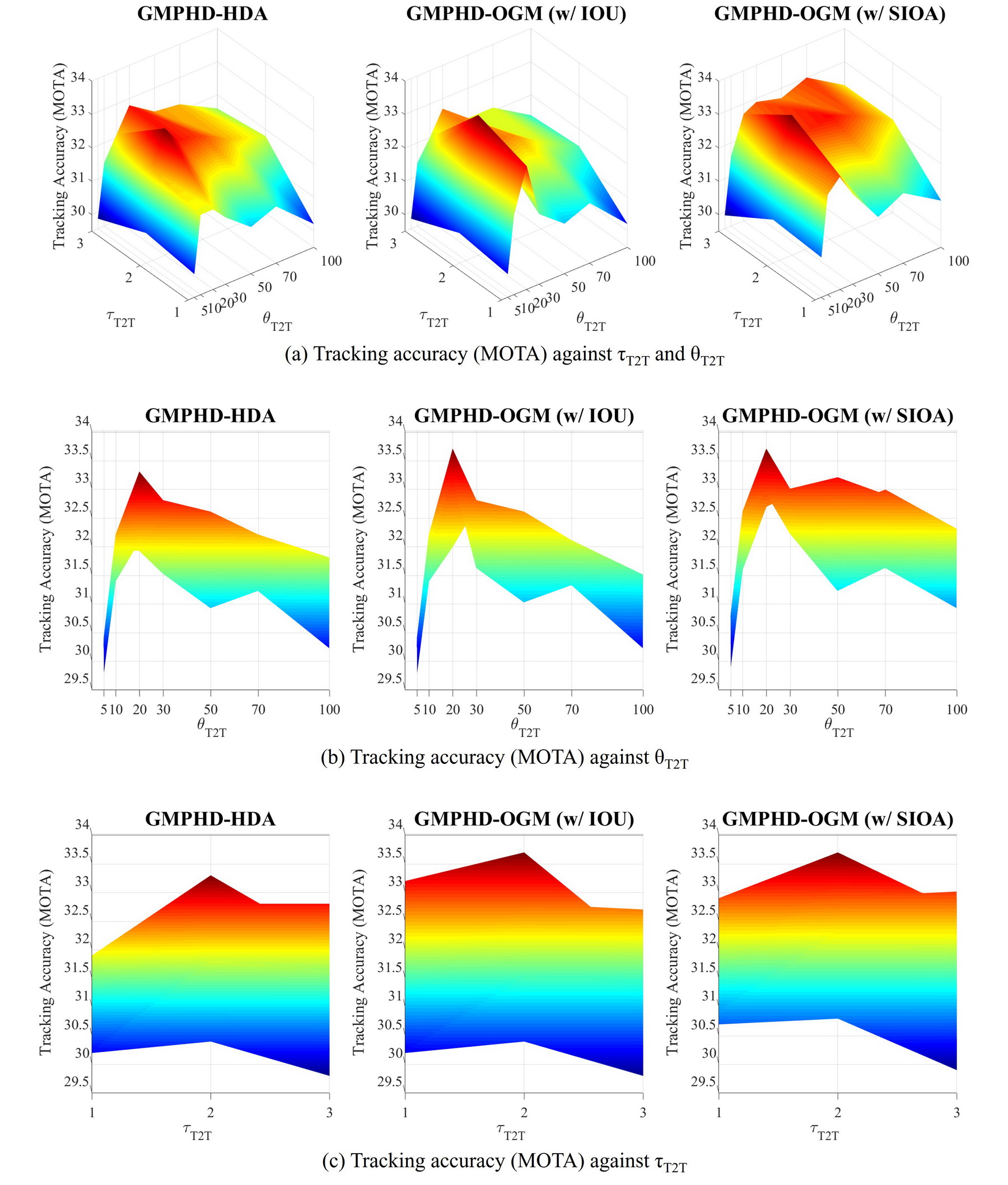}
\caption{Ablation study with the two baseline methods, i.e., GMPHD-HDA and GMPHD-OGM (with IOU) IOU. The final proposed method is GMPHD-OGM (with SIOA). Three graphs indicates the MOTA scores' distributions against (a) the minimum track length for T2TA ($\tau_{T2T}$) and the maximum frame interval for T2TA ($\theta_{T2T}$), (b) $\tau_{T2T}$, and (c) $\theta_{T2T}$. GMPHD-OGM (with SIOA) shows overall improvement in upper and lower bound of MOTA score.}
\label{fig:ablation}
\end{figure*}

\begin{figure*}[p]
\centering
\includegraphics[width=18cm]{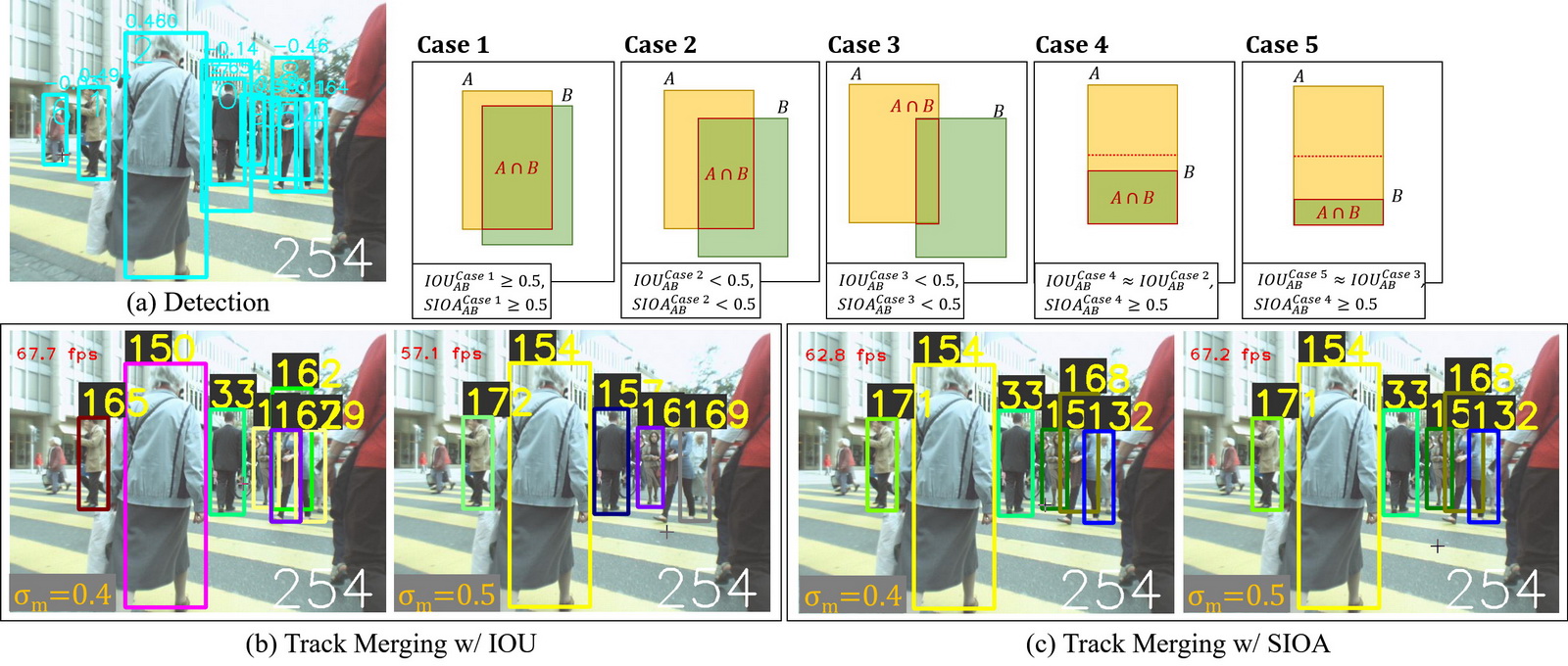}
\caption{Case study about ``Track Merging" with the qualitative results on MOT17-05-DPM training sequence at frame 254. For the same detection results, the overlapping ratios between the occluded objects are measured with $IOU\succsim0.4$ and $SIOA\succsim0.6$. Under the different merging threshold $\sigma_{m}$ values $0.4$ and $0.5$, the IOU metric is more sensitive than the SIOA metric. The SIOA metric is more robust to merge size variant false positive bounding boxes than IOU metric.}
\label{fig:merge}
\end{figure*}

First, in the ablation study, we employ the two baseline methods, to find optimal parameters settings and to prove the effectiveness of the proposed method. One is the GMPHD filter based tracker with the hierarchical data association (HDA) and without the occlusion group management (OGM). The other is the GMPHD filter based tracker with HDA and OGM by using the IOU metric for measuring occlusion ratio. We name these three methods as GMPHD-HDA, GMPHD-OGM (/w IOU), and GMPHD-OGM (/w SIOA) as shown in Table~\ref{table:eval_train_mot15} and Figure~\ref{fig:ablation}. The GMPHD-OGM (/w SIOA) method is our final tracking model. 
The scene-by-scene optimal parameter settings of those three methods are obtained by another ablation study as shown in Figure~\ref{fig:ablation}. The same $\tau_{T2T}$ and $\theta_{T2T}$ settings are applied to the whole training sequences with the range $\{1,2,3\}$ and $\{5,10,20,30,50,70,100\}$, respectively. GMPHD-OGM (/w IOU) is improved over GMPHD-HDA in terms of the upper bound of tracking accuracy (the maximum MOTA). GMPHD-OGM (/w SIOA) shows that upper bound and lower bound of tracking accuracy increases. Besides, with $\theta_{T2T}$ over $20$, the maximum and minimum values of MOTA increase on average. 
Figure~\ref{fig:merge} shows the comparison results of ``Track Merging" between using the IOU metric and the SIOA metric when the detection results with a lot of false positives are given. Also, through the case study on occlusion, we observe that the IOU metric cannot consider size-variant detections with false positives and too sensitive to be used for merging as shown in Figure~\ref{fig:merge}. On the other hand, the SIOA metric can consider the size-variant detection and the optimal value of merging threshold $\sigma_{m}$ is decided to be $0.5$ by the occlusion cases 4 and 5, empirically.
Table~\ref{table:eval_train_mot15} provides the quantitative results on the MOT15 training dataset with the best performance results on each sequence and $\sigma_{m}=0.5$. GMPH-OGM (w/ IOU) does not show outstanding improvement compared to GMPHD-HDA even though GMPHD-OGM (w/ IOU) takes more processing time since the OGM scheme runs whereas it does not in GMPHD-HDA. GMPHD-OGM (w/ SIOA) shows meaningful improvements in terms of MOTA,
The ablation study proves that GMPHD-OGM is not only overall improved but also more robust and less sensitive in parameters than baseline methods. 

\begin{figure*}[p]
\centering
\includegraphics[width=18cm]{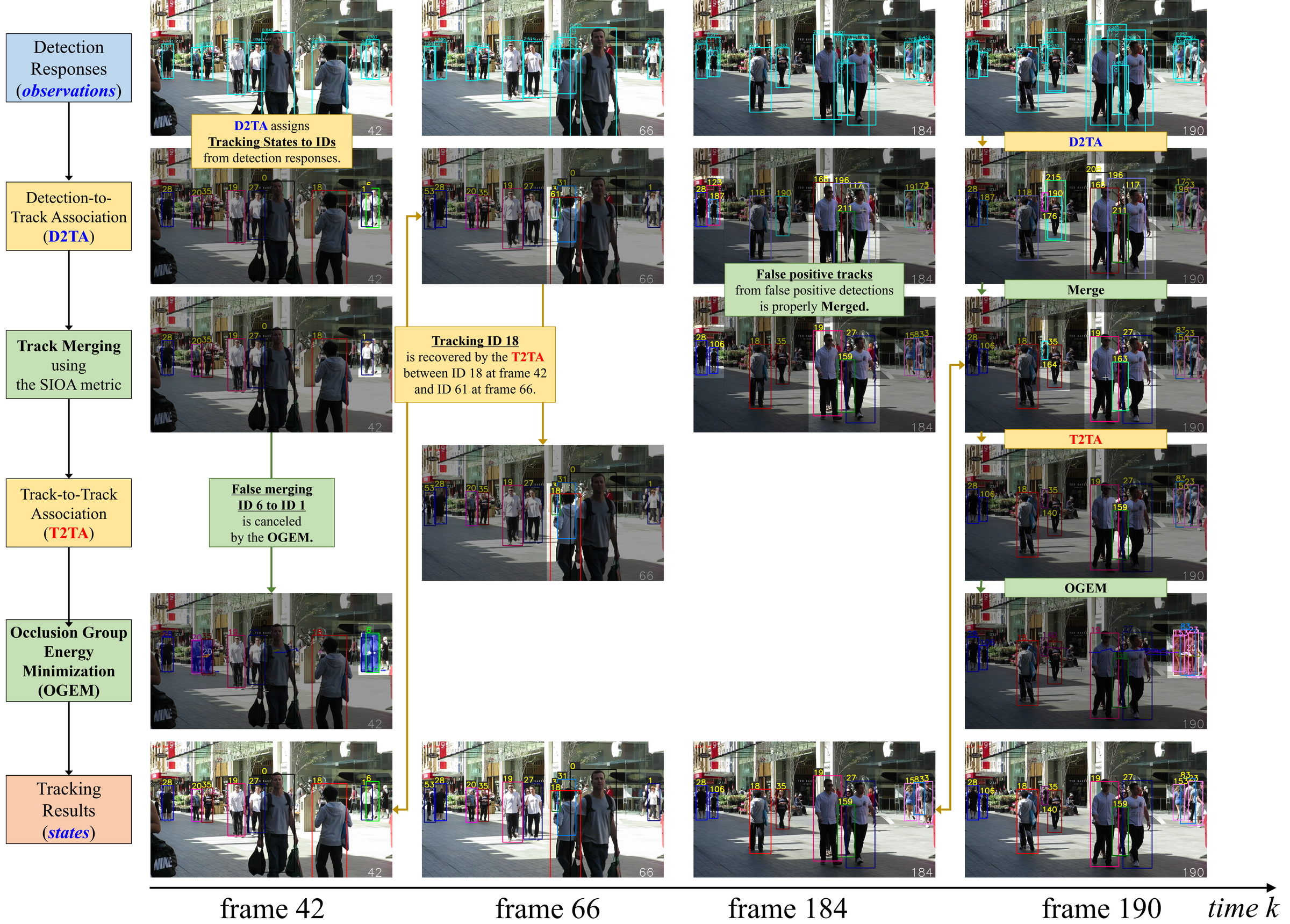}
\caption{Illustration of the proposed multi-object tracking process with the qualitative results on MOT17-08-DPM test sequence. The whole process consists of four components which are D2TA, Merge, T2TA, and OGEM. Qualitative tracking results at frame 42, 66, 184, and 190 present that all components are complementary to each other with handling tracking problems.}
\label{fig:qual_results}
\end{figure*}

Figure~\ref{fig:qual_results} demonstrates some qualitative results of our tracking framework in view of overall process.
Detection results (observations) initialize tracking objects (states).
In the sequential tracking process, the states are associated with the proper observations by the detection-to-track association (D2TA) using the GMPHD filtering process. 
From false positive detections, false positive tracks can be generated and then ``Track Merging" handles it.
If objects are occluded and their IDs are switched, track-to-track association (T2TA) can recover their IDs. 
In the case that ``Track Merging" merges true tracks (false merging), the occlusion group energy minimization (OGEM) process can recover it which optimizes energy of a group of occluded objects at current time by calculating the probability of the Gaussian mixture model as described in Subsection~\ref{subsec:ogem}.

Table~\ref{table:eval_test_mot15} and \ref{table:eval_test_mot17} show the quantitative evaluations results on MOT15 and MOT17 test dataset, respectively.
Those two benchmark datasets have crucial different characteristics.
First, provided public detection results are different, MOT15 provides ACF~\cite{acf} detector based detections, and MOT17 provides three types of detections such as DPM~\cite{dpm}, FRCNN~\cite{frcnn}, and SDP~\cite{sdp}. Compared to the DNN based detectors FRCNN and SDP, ACF and DPM exploit hand-crafted features learning and models, and thus show relatively poor performance. DPM generates more false positives than FRCNN and SDP do, and especially ACF misses much more objects than others do. Thus, in MOT15, state-of-the-art trackers shows wider range of MOTA distribution than that in MOT17. Among online methods, the trackers with DNN~\cite{bae2,motdt} shows the top MOTA scores in MOT15 and MOT17, respectively. Our method achieves the second best MOTA 30.7 vs. the best speed 169.5 fps in MOT15, but we think that the performance is competitive and enough to consider real-time application. 
\begin{figure}[t]
\centering
\includegraphics[width=9cm]{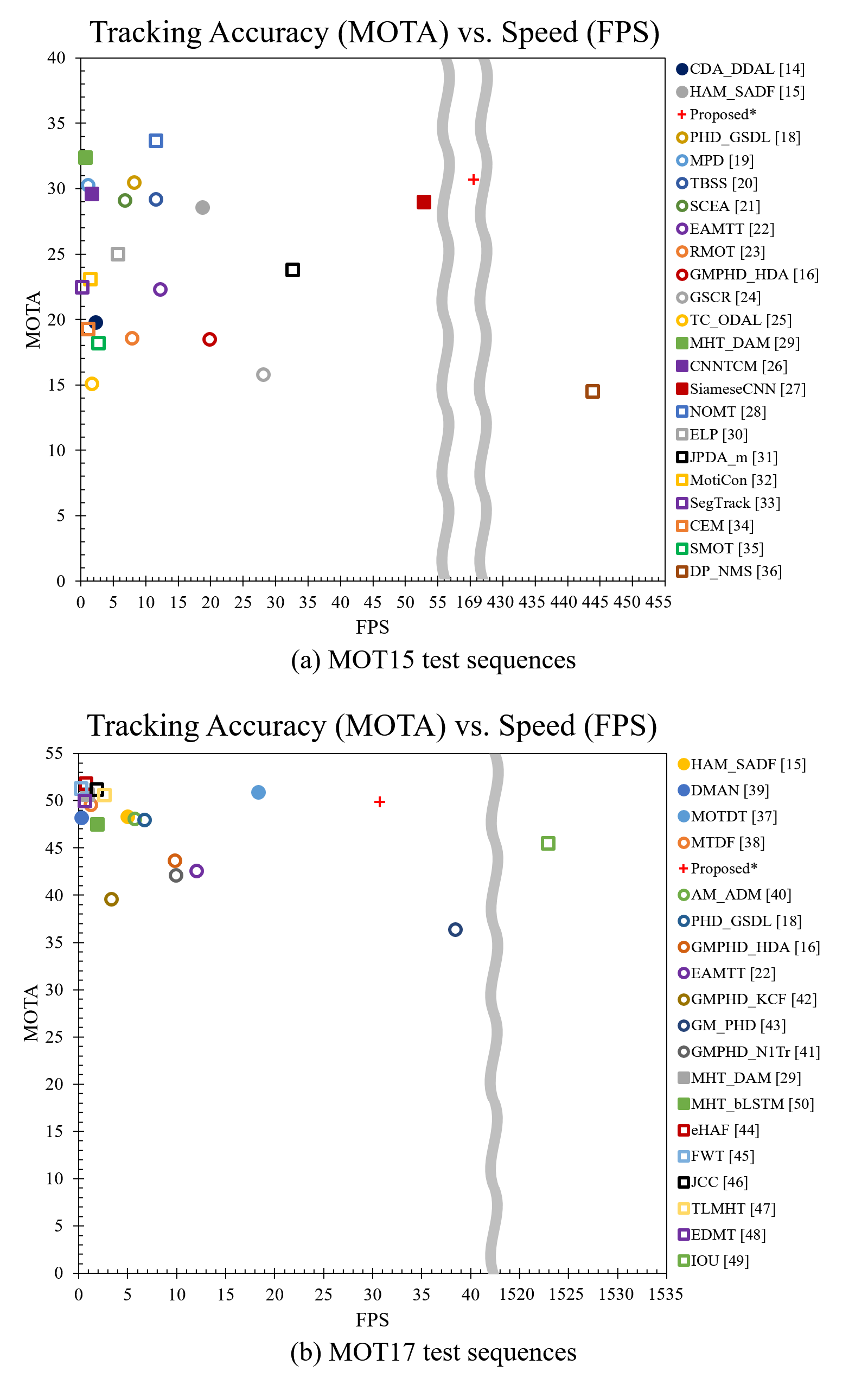}
\caption{Comparisons of tracking accuracy against speed with the state-of-the-art methods on the (a) MOT15 and (b) MOT17 test sequences.}
\label{fig:mota_fps}
\end{figure}
In Figure~\ref{fig:mota_fps}-(a), the proposed method is located in a distinguished spot in terms of tracking accuracy (MOTA) vs. speed (fps). That also proves effectiveness of our occlusion group based object analysis (OGM), compared to other relation analysis between all objects in the scene~\cite{scea,rmot}.
However, in MOT17, the speed of the proposed method decreases to 30.7 fps. That speed still belongs to real-time processing time but is not outstanding compared to other online methods.
That is caused by the second different point between two datatsets.
MOT15 includes 5783 frames with 721 tracks, 61440 bounding boxes, and 10.6 density i.e., the average number of objects a frame, whereas MOT17 includes 17757 frames with 2355 tracks, 564228 bounding boxes, and 31.8 density.
Because MOT17 has the scenes not only with much higher density but also accurate detection results, those points increase tracking accuracy and processing time. Figure~\ref{fig:mota_fps}-(b) proves those facts where the performances of state-of-the-art methods are saturated on the spot with MOTA around 50 and speed under 5 fps.
Even though our method achieves the second best MOTA and speed among online approaches in MOT17, the speed is drastically decreased compared to MOT15. 
\begin{figure}[t]
\centering
\includegraphics[width=9cm]{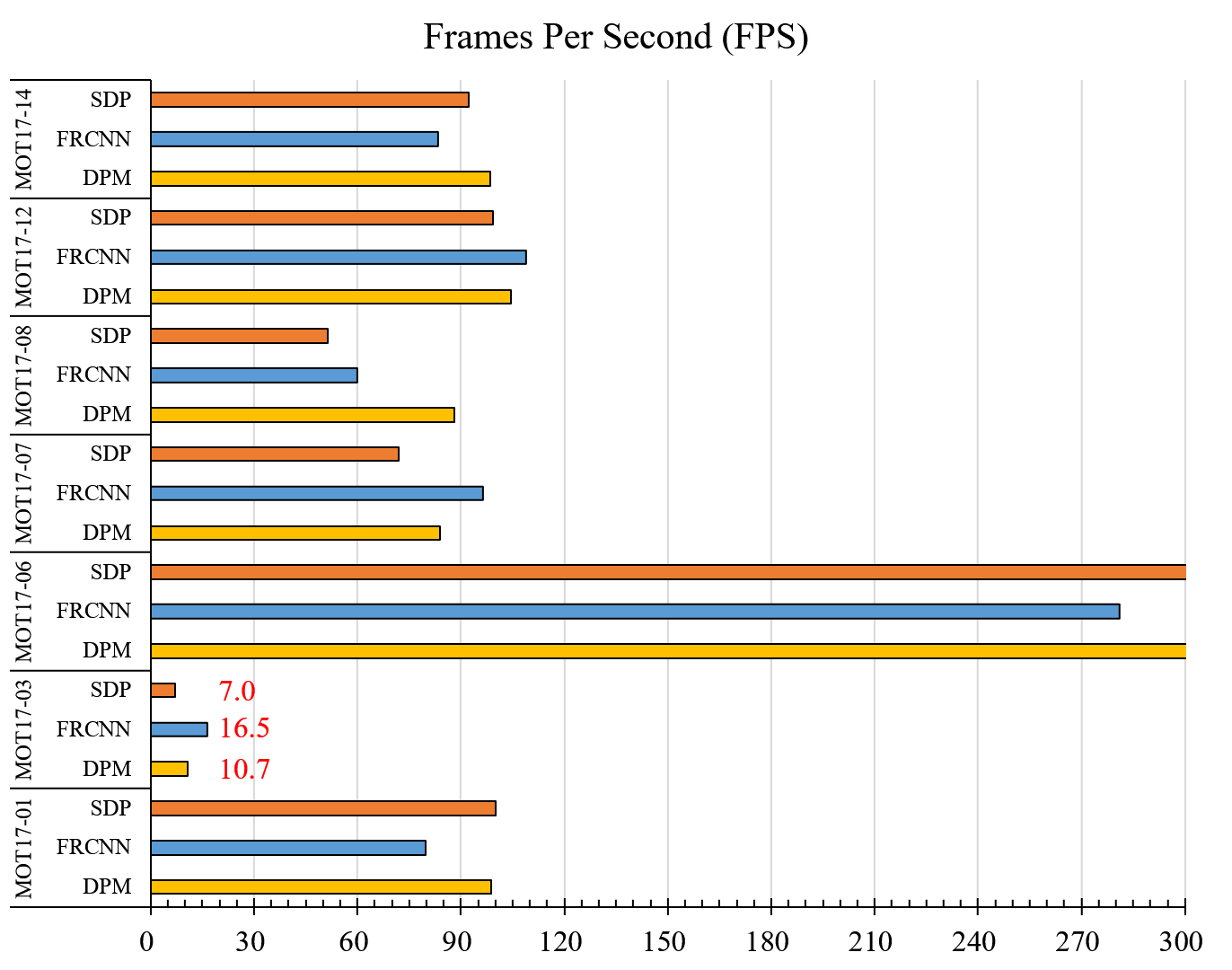}
\caption{Speed comparison of the proposed tracking method on MOT17 test dataset which provides three types of detection results for each scene, including DPM~\cite{dpm}, FRCNN~\cite{frcnn}, and SDP~\cite{sdp}.}
\label{fig:fps_scene}
\end{figure}
Figure~\ref{fig:fps_scene} explains that reason. In MOT17-03, the speed is around 10 fps since many objects appear in the scene with 69.8 density which means the average number of objects per frame. That makes the number of track-to-track association greatly increase.
The proposed method is still comparative and positioned at meaningful spot for real-time application as shown in Figure~\ref{fig:mota_fps}-(b).
In addition to our tracking algorithm (GMPHD-OGM), many PHD filter based online approaches~\cite{fu1, mtdf, prev1, eamtt, gmphdn1tr, gmphdkcf, gmphd2012} have been proposed in the past decade. Against them, GMPHD-OGM achieves not only the best MOTA, MOTP, MT, ML, FN, and speed scores on MOT15 but also the second best MOTA, speed, and best MT, FN, and Frag scores in MOT17.
Especially, against to state-of-the-art online approaches, the proposed method is distinguished in terms of tracking accuracy (MOTA) vs. speed (fps), even though we did not utilize any complex visual features except bounding boxes.
Also, the proposed tracker (GMPHD-OGM) against state-of-the-art algorithms including online with DNN even including offline, GMPHD-OGM shows the competitive MOTA versus speed as described in Figure~\ref{fig:mota_fps}, Table~\ref{table:eval_test_mot15}, and Table~\ref{table:eval_test_mot17}.

\begin{table*}[t]
\centering
\caption{Quantitative evaluation results on MOT15 training dataset. The proposed method namely GMPHD-OGM (w/ SIOA) is compared to two baseline methods GMPHD-HDA and GMPHD-OGM (w/ IOU). GMPHD-HDA employs the GMPHD filtering with hierarchical data association (HDA). GMPHD-OGM is equal to GMPHD-HDA with the proposed occlusion group management (OGM). The IOU and SIOA metrics are used for ``Track Merging" in GMPHD-OGM (w/ IOU) and (w/ SIOA), respectively. The optimal values of the merging threshold $\sigma_{m}$ are \underline{underlined} and the best scores are in \textbf{bold} in terms of the CLEAR-MOT metrics.}
\label{table:eval_train_mot15}
 \begin{tabular}{|c|c|ccccccccc|} 
        
            \hline
             \footnotesize{Tracker} & $\sigma_{m}$ & \footnotesize{MOTA$\uparrow$} &  \footnotesize{MOTP$\uparrow$} & \footnotesize{MT$\uparrow$} &  \footnotesize{ML$\downarrow$} &  \footnotesize{FP$\downarrow$} &  \footnotesize{FN$\downarrow$} &  \footnotesize{IDS$\downarrow$} & \footnotesize{Frag$\downarrow$} &  \footnotesize{Speed$\uparrow$} \\ 
             \hline\hline

           \footnotesize{GMPHD-HDA} & n/a & \footnotesize{34.8 \%} &  \footnotesize{72.3 \%} &  \footnotesize{14.4 \%} &
           \footnotesize{47.8 \%} &  
           \footnotesize{4,042} &  \footnotesize{21,646} &  \footnotesize{338} &  \footnotesize{572} &  \footnotesize{212.4 fps}  \\ 
           \hline
           
           \multirow{4}{*}{GMPHD-OGM (w/ IOU)}
            & 0.2 & \footnotesize{34.5 \%} &  \footnotesize{\textbf{72.4 \%}} &  \footnotesize{13.4 \%} &  \footnotesize{49.8 \%} &  
           \footnotesize{3,594} &  \footnotesize{22,226} &  \footnotesize{285} &  \footnotesize{550} &  \footnotesize{201.1 fps}  \\ 
           
            & \underline{0.3} & \footnotesize{35.6 \%} &  \footnotesize{72.3 \%} &  \footnotesize{14.0 \%} &  \footnotesize{48.6 \%} &  
           \footnotesize{3,537} &  \footnotesize{21,901} &  \footnotesize{\textbf{278}} &  \footnotesize{548} &  \footnotesize{\textbf{228.4 fps}}  \\ 
           
           & 0.4 & \footnotesize{35.3 \%} &  \footnotesize{\textbf{72.4 \%}} &  \footnotesize{14.0 \%} &  \footnotesize{48.2 \%} &  
           \footnotesize{3,667} &  \footnotesize{21,865} &  \footnotesize{291} &  \footnotesize{562} &  \footnotesize{201.9 fps}  \\ 
           
          & 0.5 & \footnotesize{34.7 \%} &  \footnotesize{72.3 \%} &  \footnotesize{14.4 \%} &  \footnotesize{47.8 \%} &  
           \footnotesize{4,044} &  \footnotesize{21,661} &  \footnotesize{340} &  \footnotesize{577} &  \footnotesize{205.3 fps}  \\ 
           \hline
             
            \multirow{5}{*}{GMPHD-OGM (w/ SIOA)} 
            
            & 0.3 & \footnotesize{34.5 \%} &  \footnotesize{72.3 \%} &  \footnotesize{14.2 \%} &  \footnotesize{49.2 \%} &  
           \footnotesize{\textbf{3,496}} &  \footnotesize{22,336} &  \footnotesize{297} &  \footnotesize{\textbf{559}} &  \footnotesize{216.3 fps}  \\ 
            
            & 0.4 & \footnotesize{35.4 \%} &  \footnotesize{\textbf{72.4 \%}} &  \footnotesize{14.6 \%} &  \footnotesize{48.4 \%} &  
           \footnotesize{3,556} &  \footnotesize{21,930} &  \footnotesize{284} &  \footnotesize{\textbf{540}} &  \footnotesize{216.3 fps}  \\ 

           & \underline{0.5} & \footnotesize{\textbf{35.8 \%}} &  \footnotesize{72.2 \%} &  \footnotesize{15.0 \%} &  \footnotesize{47.6 \%} &  
           \footnotesize{3,569} &  \footnotesize{21,758} &  \footnotesize{295} &  \footnotesize{545} &  \footnotesize{221.0 fps}  \\ 
           
           & 0.6 & \footnotesize{35.5 \%} &  \footnotesize{72.3 \%} &  \footnotesize{\textbf{15.6 \%}} &  \footnotesize{\textbf{47.2 \%}} &  
           \footnotesize{3,702} &  \footnotesize{21,724} &  \footnotesize{312} &  \footnotesize{556} &  \footnotesize{221.9 fps}  \\ 
           
           & 0.7 & \footnotesize{34.7 \%} &  \footnotesize{72.2 \%} &  \footnotesize{\textbf{15.6 \%}} &  \footnotesize{\textbf{47.2 \%}} &  
           \footnotesize{4,159} &  \footnotesize{\textbf{21,519}} &  \footnotesize{368} &  \footnotesize{567} &  \footnotesize{202.9 fps}  \\ 
           \hline
        \end{tabular}
\end{table*}

\begin{table*}[t]
\centering
\caption{Quantitative evaluation results on MOT15 test dataset. The proposed method is compared to state-of-the-art in terms of the CLEAR-MOT metrics. For each mode, i.e, online and offline, the first and the second best scores are highlighted in \textbf{\textcolor{red}{red}} and \textcolor{blue}{blue} in terms of each metric.}
\label{table:eval_test_mot15}
 \begin{tabular}{|c|c|c|ccccccccc|} 
        
            \hline
             \footnotesize{Mode}& \footnotesize{Tracker} & \footnotesize{DNN}& \footnotesize{MOTA$\uparrow$} &  \footnotesize{MOTP$\uparrow$} & \footnotesize{MT$\uparrow$} &  \footnotesize{ML$\downarrow$} &  \footnotesize{FP$\downarrow$} &  \footnotesize{FN$\downarrow$} &  \footnotesize{IDS$\downarrow$} & \footnotesize{Frag$\downarrow$} &  \footnotesize{Speed$\uparrow$} \\ 
             \hline\hline
             
            \multirow{12}{*}{\rotatebox[origin=c]{90}{Online}}
           & \footnotesize{CDA\_DDAL~\cite{bae2}} &  \footnotesize{O}
           & \footnotesize{\textbf{\textcolor{red}{32.8 \%}}} &  \footnotesize{70.7 \%} &  \footnotesize{9.7 \%} &  \footnotesize{42.2 \%} &  
           \footnotesize{4,983} &  \footnotesize{35,690} &  \footnotesize{614} &  \footnotesize{1,583} &  \footnotesize{2.3 fps}  \\ 
           
           & \footnotesize{HAM\_SADF~\cite{ham}} &  \footnotesize{O}
           & \footnotesize{28.6 \%} &  \footnotesize{71.1 \%} &  \footnotesize{10.0 \%} &  \footnotesize{44.0 \%} &  
           \footnotesize{7,485} &  \footnotesize{35,910} &  \footnotesize{\textcolor{blue}{460}} &  \footnotesize{\textcolor{blue}{1,038}} &  \footnotesize{18.7 fps}  \\ \cline{2-12}
             
           & \footnotesize{Proposed*} &  \footnotesize{X}
           & \footnotesize{\textcolor{blue}{30.7 \%}} &  \footnotesize{\textbf{\textcolor{red}{71.6 \%}}} &  \footnotesize{\textcolor{blue}{11.5 \%}} &  \footnotesize{\textbf{\textcolor{red}{38.1 \%}}} &  
           \footnotesize{6,502} &  \footnotesize{\textcolor{blue}{35,030}} &  \footnotesize{1,034} &  \footnotesize{1,351} &  \footnotesize{\textbf{\textcolor{red}{169.5 fps}}}  \\ 
             
            & \footnotesize{PHD\_GSDL~\cite{fu1}} &  \footnotesize{X}
            & \footnotesize{30.5 \%} &  \footnotesize{71.2 \%} &  \footnotesize{7.6 \%} &  \footnotesize{41.2 \%} &  \footnotesize{6,534} &  \footnotesize{35,284} &  \footnotesize{879} &  \footnotesize{2,208} &  \footnotesize{8.2 fps}   \\
            
            & \footnotesize{MDP~\cite{mdp}} &  \footnotesize{X}
            & \footnotesize{30.3 \%} &  \footnotesize{\textcolor{blue}{71.3 \%}} &  \footnotesize{\textbf{\textcolor{red}{14.0 \%}}} &  \footnotesize{\textcolor{blue}{38.4 \%}} &  \footnotesize{9,717} &  \footnotesize{\textbf{\textcolor{red}{32,422}}} &  \footnotesize{680} &  \footnotesize{1,500} &  \footnotesize{1.1 fps}  \\
            
           &\footnotesize{TBSS~\cite{tbss}} &  \footnotesize{X}
           & \footnotesize{29.2 \%} &  \footnotesize{\textcolor{blue}{71.3 \%}} &  \footnotesize{6.8 \%} &  \footnotesize{43.8 \%} &  \footnotesize{\textcolor{blue}{6,068}} &  \footnotesize{36,779} &  \footnotesize{649} &  \footnotesize{1,508} 
            &  \footnotesize{11.5 fps}   \\
            
            &\footnotesize{SCEA~\cite{scea}} &  \footnotesize{X}
            & \footnotesize{29.1 \%} &  \footnotesize{71.1 \%} &  \footnotesize{8.9 \%} &  \footnotesize{47.3 \%} &  \footnotesize{\textbf{\textcolor{red}{6,060}}} &  \footnotesize{36,912} &  \footnotesize{604} &  \footnotesize{1,182}
            &  \footnotesize{6.8 fps}   \\
         
            & \footnotesize{EAMTT~\cite{eamtt}} &  \footnotesize{X}
            & \footnotesize{22.3 \%} &  \footnotesize{69.6 \%} 
            &  \footnotesize{5.4 \%} &  \footnotesize{52.7 \%} &  \footnotesize{7,924} &  \footnotesize{38,982} &  \footnotesize{833} &  \footnotesize{1,485} 
            & \footnotesize{12.2 fps}   \\
         
            & \footnotesize{RMOT~\cite{rmot}} &  \footnotesize{X}
            & \footnotesize{18.6 \%} &  \footnotesize{69.6 \%} 
            &  \footnotesize{5.3 \%} &  \footnotesize{53.3 \%} &  \footnotesize{12,473} &  \footnotesize{36,835} &  \footnotesize{684} &  \footnotesize{1,282} 
            & \footnotesize{7.9 fps}   \\
           
           & \footnotesize{GMPHD\_HDA~\cite{prev1}} &  \footnotesize{X}
           & \footnotesize{18.5 \%} &  \footnotesize{70.9 \%} &  \footnotesize{3.9 \%} &  \footnotesize{55.3 \%} &  \footnotesize{7,864} &  \footnotesize{41,766} &  \footnotesize{\textbf{\textcolor{red}{459}}} &  \footnotesize{1,266} &  \footnotesize{19.8 fps}   \\
           
            & \footnotesize{GSCR~\cite{gscr}} &  \footnotesize{X}
            & \footnotesize{15.8 \%} &  \footnotesize{69.4 \%} &  \footnotesize{1.8 \%} &  \footnotesize{61.0 \%} &  \footnotesize{7,597} &  \footnotesize{43,633} &  \footnotesize{514} &  \footnotesize{\textbf{\textcolor{red}{1,010}}} &  \footnotesize{\textcolor{blue}{28.1 fps}}   \\
           
            & \footnotesize{TC\_ODAL~\cite{bae1}}&  \footnotesize{X}
            &  \footnotesize{15.1 \%} &  \footnotesize{70.5 \%} &  \footnotesize{3.2 \%} &  \footnotesize{55.8 \%} &  \footnotesize{12,970} &  \footnotesize{38,538} &  \footnotesize{637} &  \footnotesize{1,716} &  \footnotesize{1.7 fps}   \\
            
            \hline\hline
           
            \multirow{11}{*}{\rotatebox[origin=c]{90}{Offline}}
            
            & \footnotesize{MHT\_DAM~\cite{mhtdam}} &  \footnotesize{O}
            & \footnotesize{\textcolor{blue}{32.4 \%}} &  \footnotesize{\textcolor{blue}{71.8 \%}} &  \footnotesize{\textbf{\textcolor{red}{16.0 \%}}} &  \footnotesize{\textcolor{blue}{43.8 \%}} &  \footnotesize{9,064} &  \footnotesize{\textbf{\textcolor{red}{32,060}}} &  \footnotesize{\textcolor{blue}{435}} &  \footnotesize{826} &  \footnotesize{0.7 fps}  \\
            
            & \footnotesize{CNNTCM~\cite{cnntcm}} &  \footnotesize{O}
            & \footnotesize{29.6 \%} &  \footnotesize{\textcolor{blue}{71.8 \%}} &  \footnotesize{11.2 \%} &  \footnotesize{44.0 \%} &  
           \footnotesize{7,786} &  \footnotesize{34,733} &  \footnotesize{712} &  \footnotesize{943} &  \footnotesize{1.7 fps}  \\ 
           
           & \footnotesize{SiameseCNN~\cite{siameseCNN}} &  \footnotesize{O}
           & \footnotesize{29.0 \%} &  \footnotesize{71.2 \%} &  \footnotesize{8.5 \%} &  \footnotesize{48.4 \%} &  
           \footnotesize{\textbf{\textcolor{red}{5,160}}} &  \footnotesize{37,798} &  \footnotesize{639} &  \footnotesize{1,316} &  \footnotesize{\textcolor{blue}{52.8 fps}}  \\ \cline{2-12}
            
             & \footnotesize{NOMT~\cite{nomt}} &  \footnotesize{X}
             & \footnotesize{\textbf{\textcolor{red}{33.7 \%}}} &  \footnotesize{\textbf{\textcolor{red}{71.9 \%}}} &  \footnotesize{\textcolor{blue}{12.2 \%}} &  \footnotesize{44.6 \%} &  \footnotesize{7,762} &  \footnotesize{\textcolor{blue}{32,547}} &  \footnotesize{442} &  \footnotesize{\textcolor{blue}{823}} &  \footnotesize{11.5 fps}  \\
            
            & \footnotesize{ELP~\cite{elp}} &  \footnotesize{X}
            & \footnotesize{25.0 \%} &  \footnotesize{71.2 \%} &  \footnotesize{7.5 \%} &  \footnotesize{\textcolor{blue}{43.8 \%}} &  \footnotesize{7,345} &  \footnotesize{37,344} &  \footnotesize{1,396} &  \footnotesize{1,804} &  \footnotesize{5.7 fps}  \\
            
            &\footnotesize{JPDA\_m~\cite{jpdam}} &  \footnotesize{X}
            & \footnotesize{23.8 \%} &  \footnotesize{68.2 \%} &  \footnotesize{5.0 \%} &  \footnotesize{58.1 \%} &  \footnotesize{\textcolor{blue}{6,373}} &  \footnotesize{70,084} &  \footnotesize{\textbf{\textcolor{red}{365}}} &  \footnotesize{869} &  \footnotesize{32.6 fps}  \\
            
             &\footnotesize{MotiCon~\cite{moticon}} &  \footnotesize{X}
             & \footnotesize{23.1 \%} &  \footnotesize{70.9 \%} &  \footnotesize{4.7 \%} &  \footnotesize{52.0 \%} &  \footnotesize{10,404} &  \footnotesize{35,844} &  \footnotesize{1,018} &  \footnotesize{1,061} &  \footnotesize{1.4 fps}  \\
            
            &\footnotesize{SegTrack~\cite{segtrack}} &  \footnotesize{X}
            & \footnotesize{22.5 \%} &  \footnotesize{71.7 \%} &  \footnotesize{5.8 \%} &  \footnotesize{63.9 \%} &  \footnotesize{7,890} &  \footnotesize{39,020} &  \footnotesize{697} &  \footnotesize{\textbf{\textcolor{red}{737}}} &  \footnotesize{0.2 fps}  \\
            
            & \footnotesize{CEM~\cite{cem}} &  \footnotesize{X}
            & \footnotesize{19.3 \%} &  \footnotesize{70.7 \%} &  \footnotesize{8.5 \%} &  \footnotesize{46.5 \%} &  \footnotesize{14,180} &  \footnotesize{34,591} &  \footnotesize{813} &  \footnotesize{1,023} &  \footnotesize{1.1 fps}  \\
            
            & \footnotesize{SMOT~\cite{smot}} &  \footnotesize{X}
            & \footnotesize{18.2 \%} &  \footnotesize{71.2 \%} &  \footnotesize{2.8 \%} &  \footnotesize{54.8 \%} &  \footnotesize{8,780} &  \footnotesize{40,310} &  \footnotesize{1,148} &  \footnotesize{2,132} &  \footnotesize{2.7 fps}  \\
            
            & \footnotesize{DP\_NMS~\cite{dpnms}} &  \footnotesize{X}
            & \footnotesize{14.5 \%} &  \footnotesize{70.8 \%} &  \footnotesize{5.0 \%} &  \footnotesize{\textbf{\textcolor{red}{40.8 \%}}} &  \footnotesize{13,171} &  \footnotesize{34,814} &  \footnotesize{4,537} &  \footnotesize{3,090} &  \footnotesize{\textbf{\textcolor{red}{444.8 fps}}}  \\
            
            \hline
            \multicolumn{12}{l}{* The final proposed model is GMPHD-OGM (w/ SIOA).} \\
        \end{tabular}

\end{table*}

\begin{table*}[t]
\centering
\caption{Quantitative evaluation results on MOT17 test dataset. The proposed method is compared to state-of-the-art in terms of the CLEAR-MOT metrics. For each mode, i.e, online and offline, the first and the second best scores are highlighted in \textbf{\textcolor{red}{red}} and \textcolor{blue}{blue} in terms of each metric.}
\label{table:eval_test_mot17}
 \begin{tabular}{|c|c|c|ccccccccc|} 
        
            \hline
             \footnotesize{Mode}& \footnotesize{Tracker}& \footnotesize{DNN} & \footnotesize{MOTA$\uparrow$} &  \footnotesize{MOTP$\uparrow$} & \footnotesize{MT$\uparrow$} &  \footnotesize{ML$\downarrow$} &  \footnotesize{FP$\downarrow$} &  \footnotesize{FN$\downarrow$} &  \footnotesize{IDS$\downarrow$} & \footnotesize{Frag$\downarrow$} &  \footnotesize{Speed$\uparrow$} \\ 
             \hline\hline
             
            \multirow{12}{*}{\rotatebox[origin=c]{90}{Online}}
            
             & \footnotesize{MOTDT~\cite{motdt}} & \footnotesize{O}
             & \footnotesize{\textbf{\textcolor{red}{50.9 \%}}} &  \footnotesize{76.6 \%} &  \footnotesize{17.5 \%} &  \footnotesize{35.7 \%} &  \footnotesize{24,069} &  \footnotesize{\textcolor{blue}{250,768}} &  \footnotesize{2,474} &  \footnotesize{5,317} &  \footnotesize{18.3 fps}  \\
   
            & \footnotesize{HAM\_SADF~\cite{ham}} & \footnotesize{O}
            & \footnotesize{48.3 \%} & \footnotesize{\textcolor{blue}{77.2 \%}} &  \footnotesize{17.1 \%} &  \footnotesize{41.7 \%} &  \footnotesize{20,967} &  \footnotesize{269,038} &  \footnotesize{\textbf{\textcolor{red}{1,871}}} &  \footnotesize{\textbf{\textcolor{red}{3,020}}} &  \footnotesize{5.0 fps}   \\
           
            & \footnotesize{DMAN~\cite{dman}} & \footnotesize{O}
            &  \footnotesize{48.2 \%} &  \footnotesize{75.7 \%} &  \footnotesize{\textcolor{blue}{19.3 \%}} &  \footnotesize{38.3 \%} &  \footnotesize{26,218} &  \footnotesize{263,608} &  \footnotesize{\textcolor{blue}{2,194}} &  \footnotesize{5,378} &  
            \footnotesize{0.3 fps}   \\ \cline{2-12}

            & \footnotesize{Proposed*} & \footnotesize{X}
            & \footnotesize{\textcolor{blue}{49.9 \%}} &  \footnotesize{77.0 \%} &  \footnotesize{\textbf{\textcolor{red}{19.7 \%}}} &  \footnotesize{38.0 \%} &  \footnotesize{24,024} &  \footnotesize{255,277} &  \footnotesize{3,125} &  \footnotesize{\textcolor{blue}{3,540}} &  \footnotesize{\textcolor{blue}{30.7 fps}}  \\

            & \footnotesize{MTDF~\cite{mtdf}} & \footnotesize{X}
            & \footnotesize{49.6 \%} &  \footnotesize{75.5 \%} &  \footnotesize{18.9 \%} &  \footnotesize{\textbf{\textcolor{red}{33.1 \%}}} &  \footnotesize{37,124} &  \footnotesize{\textbf{\textcolor{red}{241,768}}} &  \footnotesize{5,567} &  \footnotesize{9,260} &  \footnotesize{1.2 fps}   \\

            & \footnotesize{AM\_ADM~\cite{amadm}} & \footnotesize{X}
            & \footnotesize{48.1 \%} &  \footnotesize{76.7 \%} &  \footnotesize{13.4 \%} &  \footnotesize{37.7 \%} &  \footnotesize{25,061} &  \footnotesize{265,495} &  \footnotesize{2,214} &  \footnotesize{5,027} &  \footnotesize{5.7 fps}   \\
           
            &\footnotesize{PHD\_GSDL~\cite{fu1}} & \footnotesize{X}
            & \footnotesize{48.0 \%} &  \footnotesize{\textcolor{blue}{77.2 \%}} &  \footnotesize{17.1 \%} &  \footnotesize{\textcolor{blue}{35.6 \%}} &  \footnotesize{23,199} &  \footnotesize{265,954} &  \footnotesize{3,998} &  \footnotesize{8,886} &  \footnotesize{6.7 fps}   \\
            
            & \footnotesize{GMPHD\_HDA~\cite{prev1}} & \footnotesize{X}
            & \footnotesize{43.7 \%} &  \footnotesize{76.5 \%} &  \footnotesize{11.7 \%} &  \footnotesize{43.0 \%} &  \footnotesize{25,935} &  \footnotesize{287,758} &  \footnotesize{3,838} &  \footnotesize{5,056} &  \footnotesize{9.2 fps}   \\
            
            &\footnotesize{EAMTT~\cite{eamtt}} & \footnotesize{X}
            & \footnotesize{42.6 \%} &  \footnotesize{76.0 \%} &  \footnotesize{12.7 \%} &  \footnotesize{42.7 \%} &  \footnotesize{\textcolor{blue}{20,711}} &  \footnotesize{288,474} &  \footnotesize{4,488} &  \footnotesize{5,720} &  \footnotesize{12.0 fps}   \\
            
            & \footnotesize{GMPHD\_N1Tr~\cite{gmphdn1tr}} & \footnotesize{X}
            & \footnotesize{42.1 \%} &  \footnotesize{\textbf{\textcolor{red}{77.7 \%}}} &  \footnotesize{11.9 \%} &  \footnotesize{42.7 \%} &  \footnotesize{\textbf{\textcolor{red}{18,214}}} &  \footnotesize{297,646} &  \footnotesize{10,698} &  \footnotesize{10,864} &  \footnotesize{9.9 fps}   \\
            
            & \footnotesize{GMPHD\_KCF~\cite{gmphdkcf}} & \footnotesize{X}
            & \footnotesize{39.6 \%} &  \footnotesize{74.5 \%} &  \footnotesize{8.8 \%} &  \footnotesize{43.3 \%} &  \footnotesize{50,903} &  \footnotesize{284,228} &  \footnotesize{5,811} &  \footnotesize{7,414} &  \footnotesize{3.3 fps}   \\
            
            & \footnotesize{GM\_PHD~\cite{gmphd2012}} & \footnotesize{X}
            & \footnotesize{36.4 \%} &  \footnotesize{76.2 \%} &  \footnotesize{4.1 \%} &  \footnotesize{57.3 \%} &  \footnotesize{23,723} &  \footnotesize{330,767} &  \footnotesize{4,607} &  \footnotesize{11,317} &  \footnotesize{\textbf{\textcolor{red}{38.4 fps}}}   \\ \hline\hline
            
            \multirow{8}{*}{\rotatebox[origin=c]{90}{Offline}}
            
             & \footnotesize{MHT\_DAM~\cite{mhtdam}} & \footnotesize{O}
             & \footnotesize{50.7 \%} &  \footnotesize{\textcolor{blue}{77.5 \%}} &  \footnotesize{20.8 \%} &  \footnotesize{36.9 \%} &  \footnotesize{22,875} &  \footnotesize{252,889} &  \footnotesize{2,314} &  \footnotesize{2,865} &  \footnotesize{0.9 fps}  \\
            
             &\footnotesize{MHT\_bLSTM~\cite{mhtlstm}} & \footnotesize{O}
             & \footnotesize{47.5 \%} &  \footnotesize{\textcolor{blue}{77.5 \%}} &  \footnotesize{18.2 \%} &  \footnotesize{41.7 \%} &  \footnotesize{25,981} &  \footnotesize{268,042} &  \footnotesize{2,069} &  \footnotesize{3,124} &  \footnotesize{1.9 fps}  \\ \cline{2-12}
            
            & \footnotesize{eHAF~\cite{ehaf}} & \footnotesize{X}
            & \footnotesize{\textbf{\textcolor{red}{51.8 \%}}} &  \footnotesize{77.0 \%} &  \footnotesize{\textbf{\textcolor{red}{23.4 \%}}} &  \footnotesize{37.9 \%} &  \footnotesize{33,212} &  \footnotesize{\textbf{\textcolor{red}{236,772}}} &  \footnotesize{1,834} &  \footnotesize{\textcolor{blue}{2,739}} &  \footnotesize{0.7 fps}   \\
            
            & \footnotesize{FWT~\cite{fwt}} & \footnotesize{X}
            & \footnotesize{\textcolor{blue}{51.3 \%}} &  \footnotesize{77.0 \%} &  \footnotesize{21.4 \%} &  \footnotesize{\textbf{\textcolor{red}{35.2 \%}}} &  \footnotesize{24,101} &  \footnotesize{247,921} &  \footnotesize{2,648} &   \footnotesize{4,279} & \footnotesize{0.2 fps}   \\
          
           & \footnotesize{JCC~\cite{jcc}} & \footnotesize{X}
           & \footnotesize{51.2 \%} &  \footnotesize{75.9 \%} &  \footnotesize{20.9 \%} &  \footnotesize{37.0 \%} &  \footnotesize{25,937} &  \footnotesize{247,822} &  \footnotesize{\textcolor{blue}{1,802}} &  \footnotesize{2,984} &  \footnotesize{1.8 fps}  \\

            & \footnotesize{TLMHT~\cite{tlmht}} & \footnotesize{X}
            & \footnotesize{50.6 \%} &  \footnotesize{\textbf{\textcolor{red}{77.6 \%}}} &  \footnotesize{17.6 \%} &  \footnotesize{43.4 \%} &  \footnotesize{\textcolor{blue}{22,213}} &  \footnotesize{255,030} &  \footnotesize{\textbf{\textcolor{red}{1,407}}} &  \footnotesize{\textbf{\textcolor{red}{2,079}}} &  \footnotesize{\textcolor{blue}{2.6 fps}}  \\
            
            & \footnotesize{EDMT~\cite{edmt}} & \footnotesize{X}
            & \footnotesize{50.0 \%} &  \footnotesize{77.3 \%} &  \footnotesize{\textcolor{blue}{21.6 \%}} &  \footnotesize{\textcolor{blue}{36.3 \%}} &  \footnotesize{32,279} &  \footnotesize{\textcolor{blue}{247,297}} &  \footnotesize{2,264} &  \footnotesize{3,260} &  \footnotesize{0.6 fps}  \\
            
             &\footnotesize{IOU~\cite{iou}} & \footnotesize{X}
             & \footnotesize{45.5 \%} &  \footnotesize{76.9 \%} &  \footnotesize{15.7 \%} &  \footnotesize{40.5 \%} &  \footnotesize{\textbf{\textcolor{red}{19,993}}} &  \footnotesize{281,643} &  \footnotesize{5,988} &  \footnotesize{7,404} &  \footnotesize{\textbf{\textcolor{red}{1,522.9 fps}}}  \\
            
            \hline
            \multicolumn{12}{l}{* The final proposed model is GMPHD-OGM (w/ SIOA).} \\
        \end{tabular}
\end{table*}

\section{Conclusion and Future Work}
\label{Conclusion}
In this paper, we proposed an efficient online multi-object tracking framework with the GMPHD filter and the occlusion group management (OGM) named as GMPHD-OGM. 
In the proposed framework, our first contribution is that the Gaussian mixture probability hypothesis density (GMPHD) filter~\cite{gmphd} is exploited to resolve MOT task. Since the GMPHD filter is originally designed to handle MOT in radar/sonar system, we should revise the filter to fit to video data system. To resolve missed tracks problem in the difference domain, we extended the conventional GMPHD filtering process with the hierarchical data association (HDA) strategy as explained in Figure~\ref{fig:domains}.
The second contribution is that to solve the occlusion problems, we proposed an occlusion group management (OGM) scheme. OGM is composed of ``Track Merging" and ``Occlusion Group Energy Minimization (OGEM)". ``Track Merging" reduced the number of false positives by merging them. The OGEM prevents false merging between true tracks. Both modules complement each other, and also instead of the IOU metric, we designed a new metric named as sum-of-intersection-over-area (SIOA) to measure the occlusion ratio between visual objects.
The third is that the effectiveness of our tracker (GMPHD-OGM) was introduced by the ablation study with the baselines and the evaluation results on MOT15~\cite{MOT15} and MOT17~\cite{MOT16} benchmarks with state-of-the-art MOT methods. 
The ablation study proves that GMPHD-OGM (w/ SIOA) is more efficient to solve the defined problems than the given baseline methods such as GMPHD-HDA and GMPHD-OGM (w/ IOU).
GMPHD-OGM achieves the best MOTA scores in MOT15 and MOT17 datasets, respectively, in comparison with the PHD filter based online trackers~\cite{fu1, mtdf, prev1, eamtt, gmphdn1tr, gmphdkcf, gmphd2012}.
Finally, by the comprehensive evaluation, we conclude the proposed tracker (GMPHD-OGM) against state-of-the-art algorithms including online with DNN even including offline, GMPHD-OGM shows the competitive value in ``MOTA versus speed".
As a future work, we will develop an efficient real-time tracker even with the number of objects over hundred, simultaneously, achieving the state-of-the-art level tracking accuracy.


%





\ifCLASSOPTIONcaptionsoff
  \newpage
\fi

\vfill


\end{document}